%% file: main.tex
\documentclass[10pt,twocolumn,letterpaper]{article}

\usepackage{iccv}              

\input{ICLR2025/math_commands.tex}

\definecolor{iccvblue}{rgb}{0.21,0.49,0.74}
\usepackage[pagebackref,breaklinks,colorlinks,allcolors=iccvblue]{hyperref}
\usepackage{multirow}

\def\paperID{3902} 
\def\confName{ICCV}
\def\confYear{2025}

\title{
  \makebox[\textwidth][c]{
    \parbox[c]{1\textwidth}{
      \centering
      \bfseries
      \includegraphics[height=0.5cm]{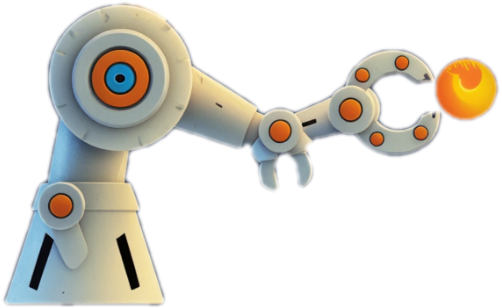} Dita: Scaling Diffusion Transformer \\for Generalist Vision-Language-Action Policy
    }
  }
}

\author{Zhi Hou$^1$\footnotemark[1] \ \ \ \ Tianyi Zhang$^{2,1}$\footnotemark[1] \ \ \ \ Yuwen Xiong$^{1}$ \ \ \ \ Haonan Duan$^{5}$\ \ \ \  Hengjun Pu$^{3,1}$ \ \ \ \ Ronglei Tong$^{5}$  \\ 
Chengyang Zhao$^{4,1}$ \ \ \ \ Xizhou Zhu$^{6, 1}$\ \ \ \ Yu Qiao$^{1}$ \ \ \ \ Jifeng Dai$^{6,1}$ \ \ \ \ Yuntao Chen$^7$\footnotemark[2] \\
$^1$ Shanghai AI Lab \ \ \ \
$^2$ 
College of Computer Science and Technology, Zhejiang University \\
$^3$ MMLab, The Chinese University of Hong Kong  \
$^4$ Peking University  \  \
$^5$ SenseTime Research \\
$^6$ Tsinghua University \ \ 
$^7$ HKISI, CAS \\
	{\small \url{https://robodita.github.io}} \\
}

\begin{document}


\makeatletter
\g@addto@macro\@maketitle{
\vspace{-3mm}
  \includegraphics[width=1.0\textwidth, trim=0 0 0 0, clip]
  {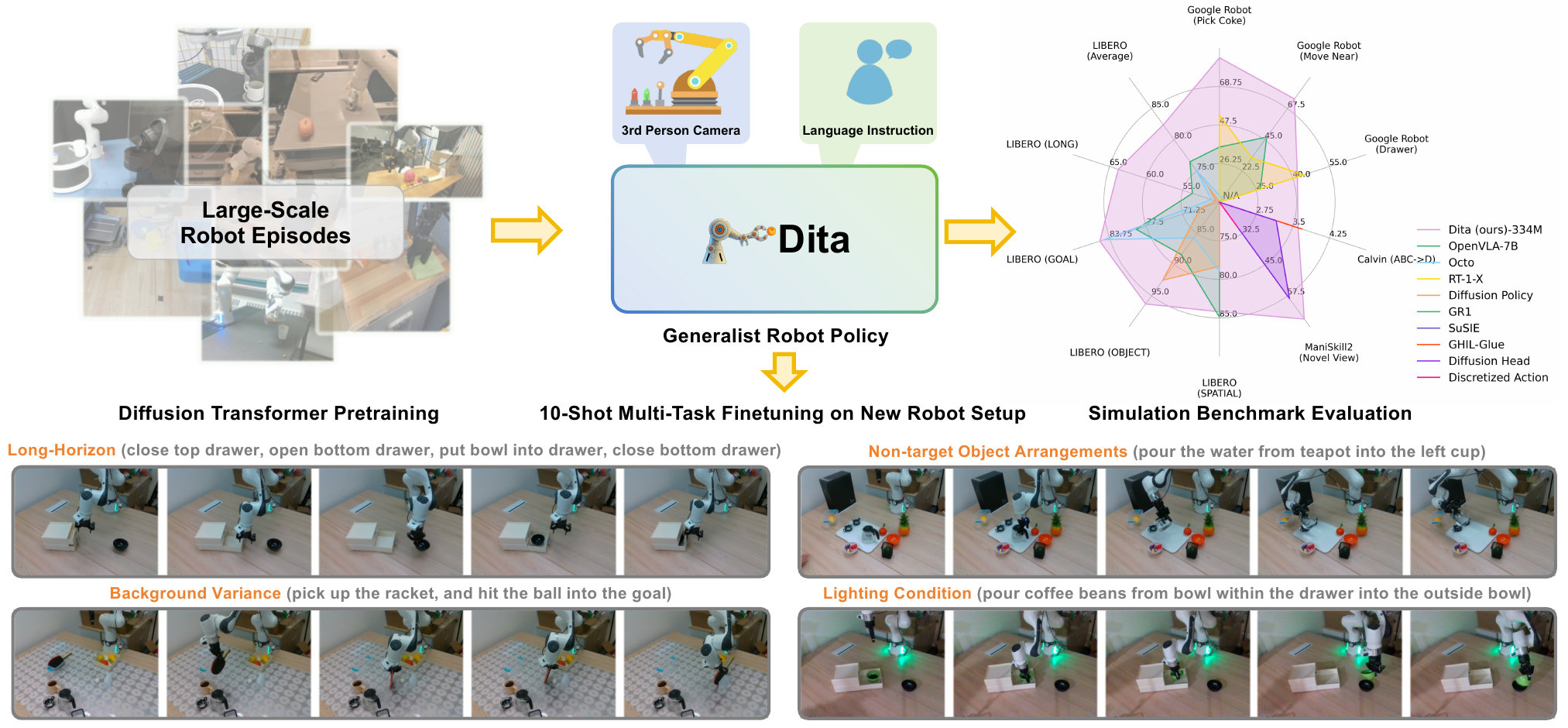}
  \vspace{-3mm}
  \captionof{figure}{We introduce Dita, an open-source, simple yet effective policy for generalist robotic learning. Pretrained on large-scale cross-embodiment datasets, Dita enables 10-shot adaptation to complex, multitask, long-horizon scenarios in novel robot setups. \textit{Particularly, Dita can complete intricate, extended-horizon tasks such as, ``close the top drawer, then open the bottom drawer, subsequently place the bowl into the bottom drawer, and finally close the bottom drawer''. Furthermore, Dita demonstrates remarkable robustness against complex object arrangements and even challenging lighting conditions in sophisticated 3D pick-and-rotation tasks.} In this context, the long-horizon demonstration scene serves as the training environment for all tasks. Additionally, Dita seamlessly scales to a wide range of popular simulation benchmarks, achieving state-of-the-art performance across these tasks.}

  \label{fig: teaser}
  \vspace{5mm}
}


\makeatother
\maketitle
\renewcommand{\thefootnote}{\fnsymbol{footnote}}
\footnotetext[1]{Equal Contribution}
\footnotetext[2]{Corresponding Author}


\begin{abstract}


While recent vision-language-action models trained on diverse robot datasets exhibit promising generalization capabilities with limited in-domain data, their reliance on compact action heads to predict discretized or continuous actions constrains adaptability to heterogeneous action spaces. We present Dita, a scalable framework that leverages Transformer architectures to directly denoise continuous action sequences through a unified multimodal diffusion process. Departing from prior methods that condition denoising on fused embeddings via shallow networks, Dita employs in-context conditioning---enabling fine-grained alignment between denoised actions and raw visual tokens from historical observations. This design explicitly models action deltas and environmental nuances. By scaling the diffusion action denoiser alongside the Transformer's scalability, Dita effectively integrates cross-embodiment datasets across diverse camera perspectives, observation scenes, tasks, and action spaces. Such synergy enhances robustness against various variances and facilitates the successful execution of long-horizon tasks. Evaluations across extensive benchmarks demonstrate state-of-the-art or comparative performance in simulation. Notably, Dita achieves robust real-world adaptation to environmental variances and complex long-horizon tasks through 10-shot finetuning, using only third-person camera inputs. The architecture establishes a versatile, lightweight and open-source baseline for generalist robot policy learning.



 





\end{abstract}

\section{Introduction}


Conventional robot learning paradigms typically depend on large-scale data collected for specific robots and tasks, yet the acquisition of data for generalized tasks remains both time-intensive and costly due to the inherent limitations of real-world robot hardware. Presently, foundational models in Natural Language Processing and Computer Vision~\cite{openai_chatgpt,openai_gpt4v,openai_dall_e,rombach2021stablediffusion,liu2023llava,chen2024internvl}, pretrained on extensive, diverse, and task-agnostic datasets, have demonstrated remarkable efficacy in addressing downstream tasks either via zero-shot approaches or with minimal task-specific samples. This achievement implies that a universal robotic policy, pretrained on heterogeneous robotic data and finetuned with minimal supervision, could be instrumental in realizing true generalization in the development of vision-language-action (VLA) models. Nevertheless, training such policies across expansive cross-embodiment datasets, encompassing diverse sensors, action spaces, tasks, camera views, and environments, remains an open challenge.

In pursuit of a unified robotic policy, recent studies have directly mapped visual observations and language instructions to actions using expansive VLA models for navigation~\cite{shah2023gnm,shah2023vint} or manipulation~\cite{brohan2022rt,brohan2023rt,kim2024openvla,team2024octo}, thereby demonstrating zero-shot or few-shot generalization in novel environments. Robot Transformers~\cite{brohan2022rt,brohan2023rt,padalkar2023open} present policy frameworks based on Transformer architectures, achieving robust generalization by training on the extensive Open X-Embodiment (OXE) Dataset~\cite{padalkar2023open}. Furthermore, Octo~\cite{team2024octo} adopts an autoregressive Transformer design with a diffusion action head, while OpenVLA~\cite{kim2024openvla} discretizes the action space and leverages a pretrained visual-language model to construct a VLA model exposed to the OXE Dataset~\cite{padalkar2023open}. Nonetheless, despite the promising potential of these VLA models~\cite{team2024octo,kim2024openvla} to learn robot policies from vast cross-embodiment datasets~\cite{padalkar2023open}, the intrinsic diversity of robot configurations within these datasets continues to constrain generalization.

\begin{figure}
    \centering
    \includegraphics[width=\linewidth]{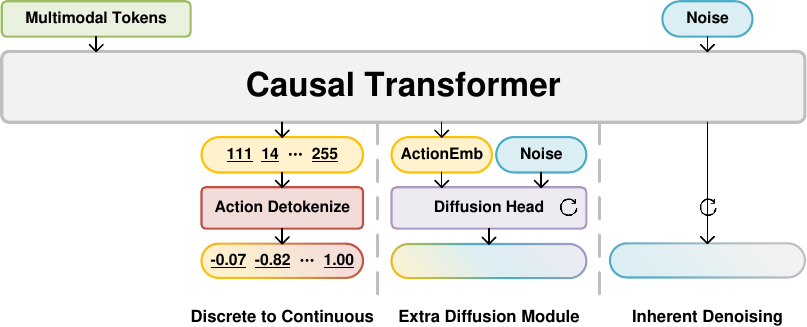}
    \caption{Illustrations of different generalist robot policy architectures. Left head: the common robot Transformer architecture with discretization actions, \eg, Robot Transformer~\cite{brohan2022rt,brohan2023rt} and OpenVLA~\cite{kim2024openvla}. Middle head: the Transformer architecture with diffusion action head which denoises the individual continuous action with a small network condition on each embedding from the causal Transformer, \eg, Octo~\cite{team2024octo} and $\pi_0$~\cite{black2024pi_0}. Right head: the proposed Dita architecture that denoises actions inherently in an in-context conditioning style.}
    %
    
    \label{fig:enter-label}
\end{figure}


Diffusion policies~\cite{chi2023diffusion,ze20243d,ke20243d,reuss2024multimodal} have demonstrated reliable performance in robotic policy learning under the paradigm of single-task imitation learning. Specifically, ~\cite{team2024octo,dasari2024ingredients} introduces a generalist policy that denoises actions using a network (MLP/DiT) as diffusion head conditioned on a single embedding from an auto-regressive multimodal Transformer. However, the expansive robot space within large-scale cross-embodiment datasets, encompassing diverse camera views and action spaces, presents a substantial challenge for a tiny diffusion head to effectively denoise continuous actions. Other diffusion policies~\cite{team2024octo,reuss2024multimodal,ke20243d} attempt to integrate historical image observations and instructions into embeddings prior to the denoising process, which might limit the denoising learning. Action anticipation typically relies more on intuitive historical observations rather than on early-fused embeddings.

In this paper, we introduce Dita, a Diffusion Transformer (DiT) Policy that capitalizes on the Transformer architecture, as demonstrated in prior work~\cite{brohan2022rt,brohan2023rt,padalkar2023open,team2024octo,kim2024openvla}, thereby ensuring scalability across extensive cross-embodiment datasets. The architecture integrates an in-context conditioning mechanism with causal transformer that intrinsically denoises action sequences, thereby enabling direct conditioning of action denoising on image tokens and empowering the model to discern subtle nuances, such as action deltas, within historical visual observations. Our objective is to provide a clean, lightweight (334M parameters), and open-source baseline model for generalist robot policy learning. The model is simple yet effective, achieving state-of-the-art or competitive results on extensive simulation benchmarks, and successfully generalizing to long-horizon tasks in novel environmental configurations---characterized by variations in background, non-target object arrangements, and lighting conditions through finetuning with a mere 10-shot set of real-world samples. Remarkably, this promising performance is achieved exclusively with a single third-person camera input, while the model's inherent flexibility affords researchers the freedom to integrate additional input modalities (e.g., wrist-camera images, target image predictions, robot state, tactile feedback, etc.) for further investigation.

\section{Related Work}

{\bf Diffusion Policy}
Denoising diffusion models~\cite{ho2020denoising,rombach2022high,dhariwal2021diffusion,peebles2023scalable,videoworldsimulators2024} have demonstrated remarkable proficiency in both image generation and multi-modal robotic action modeling~\cite{liang2024skilldiffuser,wang2024one,cao2024mamba,wang2024sparse,chen2024diffusion,chi2023diffusion,ze20243d,ke20243d,reuss2024multimodal,liu2024rdt,wen2025dexvla}. Nevertheless, existing diffusion-based manipulation policies predominantly rely on U-Net architectures or shallow cross-attention networks designed for single tasks, limiting their scalability to multi-modal applications. Recent generalist models~\cite{team2024octo,wen2024diffusion} employ VLM embeddings combined with compact MLP diffusers, while others, like RDT~\cite{liu2024rdt} and~\cite{dasari2024ingredients}, utilize cross-attention Transformers or DiT decoders for bimanual manipulation. In contrast, we propose a scalable DiT with in-context conditioning, which directly processes historical observations through a causal Transformer architecture, thereby providing enhanced expressiveness and generalization capabilities for multi-modal action generation.

\noindent{\bf Generalist Robot Policies}
Language-conditioned policies~\cite{reuss2023goal,lynch2020language,ha2023scaling,myers2023goal,zhang2022language,chen2023playfusion,tian2024predictive} have gained prominence for their adaptability in real-world applications, enabling robots to interpret and execute natural language instructions. Recent advancements in generalist robot policies leverage foundation multi-modal models across both navigation~\cite{shah2023gnm,shah2023vint,yang2024pushing,sridhar2024nomad,huang2023embodied,blessing2024information} and manipulation~\cite{bousmalis2023robocat,brohan2022rt,shah2023mutex,shridhar2023perceiver,brohan2023rt,padalkar2023open,kim2024openvla,team2024octo,driess2023palm,fang2023rh20t,pearce2023imitating,reuss2023goal,lynch2020learning,xian2023unifying,bharadhwaj2024roboagent,xiao2022robotic,mees2022matters,karamcheti2023language,scheikl2024movement}, with scalable VLA models emerging as a dominant framework~\cite{brohan2022rt,brohan2023rt,belkhale2024rt,padalkar2023open,kim2024openvla,team2024octo,qu2025spatialvla}. Some approaches incorporate large-scale video backbones trained on internet-scale data~\cite{wu2023unleashing,cheang2024gr,li2025gr,huang2025enerverse} to improve temporal visual reasoning. While these methods enhance visual representation learning, our focus is on action generation, where diffusion-based models provide a more expressive alternative. Another crucial factor in generalist policies is the choice of pretrained VLM models for action generation. Unlike recent works~\cite{black2024pi_0, li2024towards} that employ PaliGemma~\cite{beyer2024paligemma} to enhance vision-language understanding, we adopt a LLaMA-style causal Transformer for policy learning. This approach is both simple and highly scalable, demonstrating effectiveness across a wide range of benchmarks. Furthermore, by aligning robot actions with language instructions and visual observations in an in-context conditional manner, our method significantly enhances generalization across diverse robotic embodiments.










\section{Method}


In this section, we describe Dita in detail. We begin by detailing the architecture of the model, which is a scalable DiT with in-context conditioning. We then define the training objective for generating multi-modal actions. Finally, we present the data and implementation specifics for the pretraining of our model.


\begin{figure*}[h]
  \centering
  \includegraphics[width=0.98\linewidth]{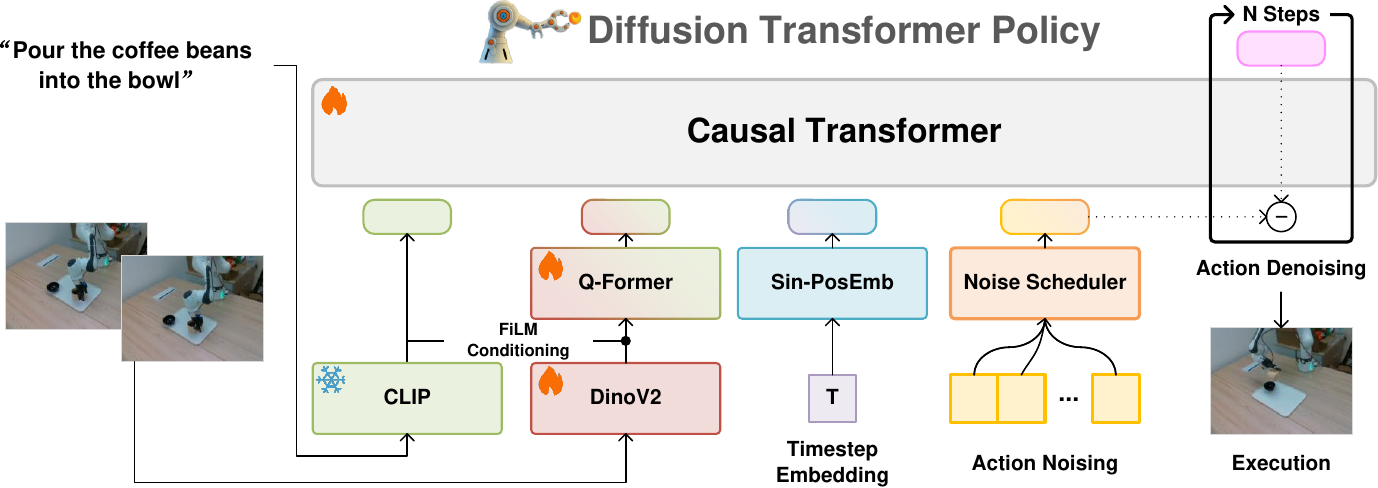}
  \caption{Our model employs a Transformer-based diffusion architecture, integrating a pretrained CLIP network to extract language instruction tokens. The DinoV2~\cite{oquab2023dinov2} model encodes image observations, followed by a Q-Former that queries features for each image. The instruction tokens, image features, timestep embeddings, and noised action are concatenated to construct a token sequence, which is then fed into the network to denoise the raw actions.}
  \label{fig:main}
\end{figure*}

\subsection{Architecture}

\noindent{\bf Multi-modal Input Tokenization}. Dita only takes language instructions and third-person camera images as input. The language instructions are tokenized using a frozen CLIP~\cite{radford2021learning} model, while the image observations are first processed by DINOv2~\cite{oquab2023dinov2} to extract image patch features. Notably, DINOv2 is trained on web data, which differs from robot-specific data. Thus, we jointly optimize the DINOv2 parameters alongside Dita in an end-to-end fashion. To mitigate computational costs, we incorporate a Q-Former~\cite{li2023blip} with FiLM~\cite{perez2018film} conditioning to select image features from the DINOv2 patch features based on the instruction context.




\noindent{\bf Action Preprocess}. We represent the end-effector action as a 7D vector, comprising 3 dimensions for the translation vector, 3 dimensions for the rotation vector, and 1 dimension for the gripper position. To align the dimensionality with the image and language tokens, we pad the continuous action vector with zeros to form the action representation. Noise is only introduced into the 7D action vector during the denoising diffusion optimization process.


\noindent{\bf Model Design}. Our core design is the DiT structure~\cite{peebles2023scalable}, which denoises the action chunk with multiple action tokens. This is achieved by conditioning directly on image observations and instruction tokens through an in-context conditioning approach using a causal Transformer. Specifically, we concatenate language tokens, image features, and timestep embeddings at the beginning of the sequence, treating the noisy action in conjunction with the instruction tokens, as illustrated in Figure~\ref{fig:main}. This design preserves the scalability of Transformer networks and enables denoising to be conditioned directly on image patches, thereby allowing the model to capture nuanced changes in action over historical observations. The model is supervised by the noise introduced into the continuous actions. In other words, we directly apply the diffusion objective in the action chunk space with a large Transformer model, in contrast to the  diffusion action head approach~\cite{team2024octo,dasari2024ingredients,reuss2024multimodal}. Notably, our proposed Dita presents a versatile and scalable design, adaptable to diverse datasets for both pretraining and finetuning, while achieving promising performance. Furthermore, additional observation tokens and input can be seamlessly integrated into the Transformer architecture. Further details are provided in Appendix A.


\subsection{Training Objective}
The denoising network $\mathcal{E}_{\theta}(c_{lang}, c_{obs}, t, \vx^t)$ is constructed upon a causal Transformer, where $c_{obs}$ represents the image observation, $c_{lang}$ denotes the language instruction, and $t \in {1,2,\dots,T_{train}}$ is the timestep index within the total denoising steps $T_{train}$. During training, a Gaussian noise vector $\vx^t \sim \mathcal{N}(\mathbf{0}, \mI)$ is sampled at each timestep $t$ and added to the action $\va$ to form the noised action token $\hat{\va}$. The network $\mathcal{E}_{\theta}$ is trained to predict the noise vector $\hat{\vx}$, with randomly sampled $t$ . The optimization objective of Dita is to minimize the mean squared error (MSE) loss between $\vx^t$ and $\hat{\vx}^t$.


The inference procedure is delineated as follows, $\alpha$, $\gamma$, and $\sigma$ constitute the noise scheduler~\cite{ho2020denoising}. The denoising process is iterated over $N_{eval}$ steps to yield a reliable action.

\begin{equation}
\vx^{t-1} = \alpha(\vx^{t} - \gamma\mathcal{E}_{\theta}(c_{lang}, c_{obs}, t, \vx^t) + \mathcal{N}(\mathbf{0}, \sigma^2\mI)).
\end{equation}

%


\subsection{Pretraining Data}

To evaluate the proposed Dita policy, we select the OXE datasets~\cite{padalkar2023open, kim2024openvla} for model pretraining. We primarily adhere to the method detailed in~\cite{team2024octo,kim2024openvla} for dataset selection and weight assignment. Actions are normalized and filtered similar to~\cite{padalkar2023open}. 




\subsection{Pretraining Details}

We employ the DDPM diffusion objective~\cite{ho2020denoising} with $T_{train}=1000$ timesteps for pretraining, while adopting DDIM~\cite{song2020denoising} with $T_{eval}=20$ timesteps during zero-shot evaluation to accelerate inference. Based on preliminary experiments reported in ManiSkill2~\cite{gu2023ManiSkill2}, we utilize 2-frame image observations to predict 16 action chunks. The network is optimized by AdamW~\cite{loshchilov2017decoupled} for 100,000 steps, with learning rates of $1e{-4}$ for both the causal Transformer and Q-Former, and $1e{-5}$ for DINOv2. Training is conducted with a batch size of 8192 across 32 NVIDIA A100 GPUs, allocating 256 samples per GPU. Additional pretraining configurations are detailed in Appendix A.





\section{Simulation Experiments}



We strive to develop a robust foundational VLA model that is both scalable across diverse simulation benchmarks and adaptive to new complex tasks in unseen robot environments with as few as 10 or even fewer samples. To assess the capabilities of the pretrained model, we conduct evaluations across four simulation benchmarks in this section: 1) SimplerEnv~\cite{li24simpler} (Google Robot) demonstrates the model's zero-shot adaptation to simulation environments; 2) LIBERO~\cite{liu2024libero} assesses finetuning adaptability with a single-camera setup; 3) Calvin~\cite{mees2022calvin} evaluates long-horizon task performance in novel environments; and 4) ManiSkill2~\cite{gu2023ManiSkill2} is re-rendered to illustrate generalization across unseen camera views. Across all four benchmarks, Dita pretrained on OXE datasets $\mathcal{E}_{\theta \sim OXE}$ is evaluated in a zero-shot manner on SimplerEnv, while it is finetuned on the remaining three benchmarks using their respective datasets.


\begin{table}
\setlength\tabcolsep{1.5pt}
\caption{Success rate comparison with RT-1-X~\cite{brohan2022rt}, Octo-Base~\cite{team2024octo} and OpenVLA-7B~\cite{kim2024openvla} on SimplerEnv (both match and variant results of Google Robot~\cite{brohan2022rt}).}
\label{tab:simplerenv}
\small
\begin{center}
\begin{tabular}{l|cc|cc|cc}
\toprule
\multirow{2}{*}{\bf Method}  &\multicolumn{2}{c}{\bf coke$\_$can} & \multicolumn{2}{c}{\bf move$\_$near } & \multicolumn{2}{c}{\bf drawer}  \\
 & { match } & {variant} & {match} & {variant} & {match} & {variant} \\
\hline 

RT-1-X~\cite{brohan2022rt} & 56.7\% & 49.0\% & 31.7\% & 32.3\% & \bf 59.7\% & 29.4\%  \\
Octo-Base~\cite{team2024octo} & 17.0\% & 0.6\% & 4.2\% & 3.1\% &  22.7\% & 1.1\%  \\
OpenVLA-7B~\cite{kim2024openvla} & 16.3\% & 54.5\% & 46.2\% & 47.7\% & 35.6\% & 17.7\% \\
Dita (Ours) & \bf 83.7\% & {\bf 85.5\%} & \bf 76.0\% & \bf 73.0\% & 46.3\% & \bf 37.5\% \\

\bottomrule
\end{tabular}

\end{center}
\end{table}

\subsection{Baselines} \label{sec:sim_baseline}


\noindent{\bf Diffusion Action Head} The diffusion head for the action generation $\mathcal{E}_{\theta \sim s}^{Diff}$~\cite{team2024octo} is also implemented. Specifically, we employ a three-layer MLP network as the denoising module, conditioned on each action token embedding outputed by the same causal Transformer architecture (as Middle head illustrated in Figure~\ref{fig:enter-label}). This approach introduces additional parameters (the extra MLP) compared to Dita.

\noindent{\bf Octo \& OpenVLA} We also reproduce these two open-source VLA models using their released checkpoints, as they employ the same multimodal inputs (language instruction and third-person camera image) as our approach.

\subsection{SimplerEnv}


SimplerEnv~\cite{li24simpler} is a Real-to-Sim platform designed to evaluate policies learned from real robot data within a simulation environment. In this section, we compare our approach with leading generalist policies, including RT-1-X~\cite{brohan2022rt,padalkar2023open}, Octo~\cite{team2024octo}, and OpenVLA~\cite{kim2024openvla}, under both match and variant scenarios. For a fair comparison, we adhere to the evaluation protocol of SimplerEnv~\cite{li24simpler}, which includes tasks ``pick up coke can'', ``move an object near to others'', ``open drawer'', ``close drawer''.

Table~\ref{tab:simplerenv} demonstrates that Dita achieves strong generalization performance under zero-shot evaluation across various types of variations, including background, texture, objects, spatial positions, and more. Leveraging the in-context conditioning design, Dita exhibits enhanced robustness, relying solely on third-person view images to detect subtle nuances and generate more reliable actions. The qualitative results are listed in Appendix~\ref{sec:simple_supp}.

\subsection{LIBERO}
\label{sec:analysis:libero}


LIBERO~\cite{liu2024libero} is a comprehensive benchmark for knowledge transfer in multitask and lifelong robot learning. It consists of four sub-datasets: LIBERO-SPATIAL, LIBERO-OBJECT, LIBERO-GOAL, and LIBERO-100. Notably, LIBERO-100 is further divided into LIBERO-90 and LIBERO-LONG, with the latter featuring 10 long-horizon tasks that encompass diverse object interactions and versatile motor skills. We employ the modified version of LIBERO from OpenVLA~\cite{kim2024openvla} as the data source for finetuning and evaluation.

\noindent{\bf Comparisons}. Table~\ref{tab:libero} demonstrates that Dita outperforms baseline methods on most LIBERO sub-datasets, achieving an overall increase in average success rate by nearly 6\%. Notably, Dita exhibits a 10\% improvement on LIBERO-LONG, highlighting its strong potential for tackling long-horizon tasks.



\subsection{CALVIN}


CALVIN~\cite{mees2022calvin} is an open-source simulated benchmark designed for learning long-horizon, language-conditioned tasks. It consists of four distinct scenes (A, B, C, and D) and introduces the ABC$\rightarrow$D evaluation protocol, where models are trained on environments A, B, and C and evaluated on environment D. The benchmark aims to solve up to 1,000 unique task sequences, each comprising five distinct subtasks. The primary evaluation metric is the success sequence length, which measures the ability to complete five consecutive subtasks within a sequence. To assess the long-horizon generalization of Dita, we adopt the ABC$\rightarrow$D setting while only utilizing static RGB images as perception inputs. Additionally, we also implement a diffusion policy baseline $\mathcal{E}_{\theta \sim s}^{Diff}$ by introducing a three-layer MLP diffusion head, further demonstrating the effectiveness of Dita’s in-context conditioning mechanism.

\begin{table}[t]
\setlength\tabcolsep{1pt}
\caption{Comparison with Diffusion Policy (denoted as DP*, training from scratch)~\cite{chi2023diffusion}, Octo~\cite{team2024octo}, and OpenVLA~\cite{kim2024openvla} on LIBERO~\cite{liu2024libero}. Except for Dita results, all other results are sourced from \cite{kim2024openvla}.}
\label{tab:libero}
\small
\begin{center}
\begin{tabular}{l|cccc|c}
\toprule
{\bf Method}  & {\bf SPATIAL} & {\bf OBJECT} & {\bf GOAL} & {\bf LONG} & {\bf Averge} \\
\hline 
DP*\cite{chi2023diffusion} & 78.3\% & 92.5\% & 68.3\% & 50.5\% & 72.4\% \\
Octo~\cite{team2024octo} & 78.9\% & 85.7\% & 84.6\% & 51.1\% & 75.1\% \\
OpenVLA~\cite{kim2024openvla} & \bf 84.9\% & 88.4\% & 79.2\% & 53.7\% & 76.5\% \\
Dita (Ours) & 84.2\% & \bf 96.3\% & \bf 85.4\% & \bf 63.8\% & \bf 82.4\% \\
\hline
Dita w/ wrist & \bf 97.4\% & 94.8\% & {\bf 93.2\%} & {\bf 83.6\%} & 92.3\% \\
\bottomrule
\end{tabular}
\end{center}

\end{table}

\begin{table*}[t]
 \centering
 \caption{The comparisons with state-of-the-art approaches on Calvin (ABC$\rightarrow$D) with the metrics of success rate and average success length. The abbreviations denote different input modalities: S-RGB for Static RGB, G-RGB for Gripper RGB, S-RGBD for Static RGB-D, G-RGBD for Gripper RGB-D, P for proprioceptive arm position, and Cam for camera parameters.}
  \label{tab:calvin}
  \small
  \begin{tabular}{c|c|ccccc|c}
\toprule    
    \multirow{2}{*}{\bf Method} & \multirow{2}{*}{\bf Input} & \multicolumn{6}{|c}{\bf No. Instructions in a Row (1000 chains)} \\ 
    \cline{3-8}
    & & 1 & 2 & 3 & 4 & 5 & Avg.Len. \\ \hline
     RoboFlamingo~\cite{li2023vision}  &S-RGB,G-RGB &	82.4\%	&61.9\%	& 46.6\%&	33.1\%	&23.5\%	& 2.47 \\

GR-1~\cite{wu2023unleashing} & S-RGB,G-RGB,P &	85.4\%	&71.2\%	 & 59.6\%	&49.7\%	& 40.1\%&	3.06 \\
3D Diffuser~\cite{ke20243d} &	S-RGBD,G-RGBD,P,Cam&	92.2\%	&78.7\%	& 63.9\%	& 51.2\%	& 41.2\%	& 3.27 \\

GR-MG~\cite{li2025gr} &	S-RGBD,G-RGBD,P&	{\bf 96.8\%}	&{\bf 
 89.3\%}	&  {\bf 81.5\%} 	& {\bf 72.7}\%	& {\bf 64.4} \%	& {\bf 4.04} \\
 \hline
SuSIE~\cite{black2023zero}&	 S-RGB &	87.0\%&	69.0\%	& 49.0\%	&38.0\%	 &26.0\%& 	2.69 \\
GHIL-Glue~\cite{hatch2024ghil,black2023zero} & S-RGB & {\bf 95.2\%}	& {\bf 88.5\%}	& {\bf 73.2\%} & {\bf 62.5\%}& {\bf 49.8\%} & {\bf 3.69} \\




\hline

$\mathcal{E}_{\theta \sim s}^{Diff}$ w/o Pretrain  & S-RGB &  75.5\%  & 44.8\%  & 25.0\% & 15.0\% & 7.5\% & 1.68 \\
$\mathcal{E}_{\theta \sim s}^{Diff}$  & S-RGB &  94.3\%  & 77.5\%  & 62.0\% & 48.3\% & 34.0\% & 3.16 \\
Ours w/o Pretrain & S-RGB &  89.5\%  & 63.3\%  & 39.8\% & 27.3\% & 18.5\% & 2.38 \\
    Ours & S-RGB &  {\bf 94.5\%}  & {\bf 82.5\%}  & {\bf 72.8\%} & {\bf 61.3\% } & {\bf 50.0\%} & {\bf 3.61} \\

%


\bottomrule
\end{tabular}

\end{table*}

\noindent{\bf Comparisons}. Table~\ref{tab:calvin} presents a comparative analysis of prior approaches and the proposed Dita on CALVIN. Without whistles and bells, the proposed Dita achieves comparable performance among methods relying solely on a single RGB camera for observation. Noticeably, only 1\% of the trajectories were labeled with text in Calvin~\cite{mees2022calvin}. The rest are unstructured play data collected by untrained users, with no information for downstream tasks. \textit{Dita does not utilize the play data which provides external trajectory data compared to the labeled data, while GR-MG uses it for training the policy.} Remarkably, GHIL-Glue~\cite{hatch2024ghil,black2023zero}, which builds upon SuSIE~\cite{black2023zero} with further finetuned generative models~\cite{brooks2023instructpix2pix,xing2024dynamicrafter}, results in significantly larger models. Furthermore, Dita surpasses its non-pretrained variant by a margin of 1.23, underscoring its superior transferability. In contrast, employing diffusion head underperforms Dita by 0.45 points with similar pretrained weights, highlighting the efficacy of Dita’s in-context conditioning mechanism. The results illustrate that Dita excels at discerning subtle visual nuances in long-horizon tasks and generalizes proficiently across diverse environments, effectively transferring knowledge from extensive, real-world pretraining datasets to the CALVIN benchmark.

\subsection{ManiSkill2}


ManiSkill2~\cite{gu2023ManiSkill2}, the next generation of the SAPIEN ManiSkill benchmark~\cite{mu2021maniskill}, serves as a widely recognized platform for assessing the generalized manipulation capabilities of embodied models. It encompasses 20 distinct manipulation task families and over 4M demonstration frames across various configurations. Leveraging ManiSkill2, we establish a \textit{novel camera view generalization} benchmark to evaluate the effectivenes of Dita.

\noindent{\bf Setup}. To construct the benchmark, we select 5 tasks (PickCube-v0, StackCube-v0, PickSingleYCB-v0, PickClutterYCB-v0, PickSingleEGAD-v0) from ManiSkill2 and create a camera pool comprising 300K random cameras. More details are provided in Appendix. We also implement RT-1~\cite{brohan2022rt} style baseline model $\mathcal{E}_{\theta \sim s}^{Disc}$ with an architecture similar to ours for comparison. Unlikely, we discretize each action dimension into 256 bins~\cite{brohan2022rt} and utilize a Transformer network to predict the corresponding bin indices.


\noindent{\bf Comparisons}. Table~\ref{tab:maniskill_main} compares the proposed method with the discretization action head and the diffusion action head. The experiments demonstrate that Dita outperforms the $\mathcal{E}_{\theta \sim s}^{Disc}$ in large-scale, novel camera view scenarios. Additionally, Dita shows superior performance on more complex tasks and outperforms $\mathcal{E}_{\theta \sim s}^{Diff}$ by 20\% in the PickSingleYCB task and by 12\% in the PickClutterYCB task. The results highlight that Dita offers better scalability on large, diverse datasets, while also achieving enhanced camera view generalization.



\begin{table}[t]
\setlength\tabcolsep{1pt}
\caption{Comparison of our model with two baseline methods (discretization and diffusion head) on ManiSkill2 success rate. The abbreviations denote the task names: S-YCB for PickSingleYCB, C-YCB for PickClutterYCB, EGAD for PickSingleEGAD.}
\label{tab:maniskill_main}
\small
\begin{center}
\begin{tabular}{l|c|ccccc}
\toprule
{\bf Method}  & {\bf Avg.} & {\bf PickC} & {\bf StackC} & {\bf S-YCB} & {\bf C-YCB} & {\bf EGAD} \\
\hline 
$\mathcal{E}_{\theta \sim s}^{Disc}$ & 30.2\% & 41.0\% & 33.0\% & 22.0\% & 1.0\% & 54.0\% \\
$\mathcal{E}_{\theta \sim s}^{Diff}$ & 58.6\% &  {\bf 86.0\%} &  76.0\% &  37.0\% &  24.0\% &  70.0\% \\
Dita (ours) & {\bf 65.8\%} &  79.0\% &  {\bf 80.0\%} &  {\bf 62.0\%} &  {\bf 36.0\%} &  {\bf 72.0\%}  \\

\bottomrule
\end{tabular}
\end{center}

\end{table}

\subsection{Ablation Study}

In this section, we conduct an ablation study on key factors in the model architecture design, including observation length, trajectory length, and denoising steps.

\noindent{\bf Observation length.} The length of historical observation images significantly impacts performance. As shown in Table~\ref{tab:maniskill_ab}, success rate drops sharply when the observation length is increased to 3. This could be due to the increased difficulty in model convergence, as the number of corresponding image tokens also rises. Additionally, we observe that using 2-frame observations enhances performance, particularly when the prediction horizon is extended. When the trajectory length is 32, Dita with 2-frame observations achieves superior performance. We argue that 2-frame observations strike an optimal balance, providing sufficient visual distinction between objects in the workspace and the robot states.

\noindent{\bf Trajectory length.} Trajectory length is the sum length of observation and action prediction chunks, which is also a critical factor influencing performance. Table~\ref{tab:maniskill_ab} shows that performance improves as trajectory length increases. Notably, the performance of more complex tasks, such as PickClutterYCB, increases substantially with longer trajectory lengths, while simpler tasks, like PickCube, maintain high performance once the trajectory length exceeds 4. Long trajectory length significantly boosts performance, as this optimization allows the model to better anticipate the target object’s position and gain awareness of more future states.



\begin{table}[t]
\setlength\tabcolsep{1pt}
\caption{Ablation on ManiSkill2 about the observation length (\# obs) and the trajectory length (\# traj).}
\label{tab:maniskill_ab}
\small
\centering
\begin{tabular}{c|c|c|ccccc}
\toprule
{\bf \# obs } & {\bf \# traj}  & {\bf All} & {\bf PickC} & {\bf StackC} & {\bf S-YCB} & {\bf C-YCB} & {\bf EGAD}\\
\hline
2 & 2 & 40.8\% &  68.0\% &  54.0\% &  33.0\% &  9.0\% &  40.0\%  \\
2 & 4 & 51.6\% &  81.0\% &  69.0\% &  44.0\% &  11.0\% &  53.0\%  \\
2 & 8 & 62.4\% &  {\bf 89.0} \% &  78.0\% &  54.0\% &  25.0\% &  66.0\%  \\
2 & 16 & 65.6\% &  83.0\% &  {\bf 80.0} \% &  {\bf 70.0} \% &  25.0\% &  70.0\% \\
2 & 32 & {\bf 65.8} \% &  79.0\% &  {\bf 80.0} \% &  62.0\% &  {\bf 36.0} \% &  {\bf 72.0}\% \\
\hline
1 & 32 & 61.6\% &  78.0\% &  76.0\% &  64.0\% &  24.0\% &  66.0\%  \\
1 & 1  & 51.0\% &  79.0\% &  66.0\% &  42.0\% &  19.0\% &  49.0\%  \\
3 & 3 & 35.4\%  &  54.0\% &  49.0\% &  27.0\% &  5.0\% &  42.0\%  \\

\bottomrule
\end{tabular}
\end{table}



\section{Real-Robot Experiments}

Few-shot real-world robot adaptation is a critical metric for evaluating the effectiveness of a generalist policy in practical applications. For real-robot experiments, We employ 10-shot finetuning to assess the model’s adaptability in complex, long-horizon, multi-modal tasks within unseen robot environments.

\begin{figure}
    \centering
    \includegraphics[width=0.95\linewidth]{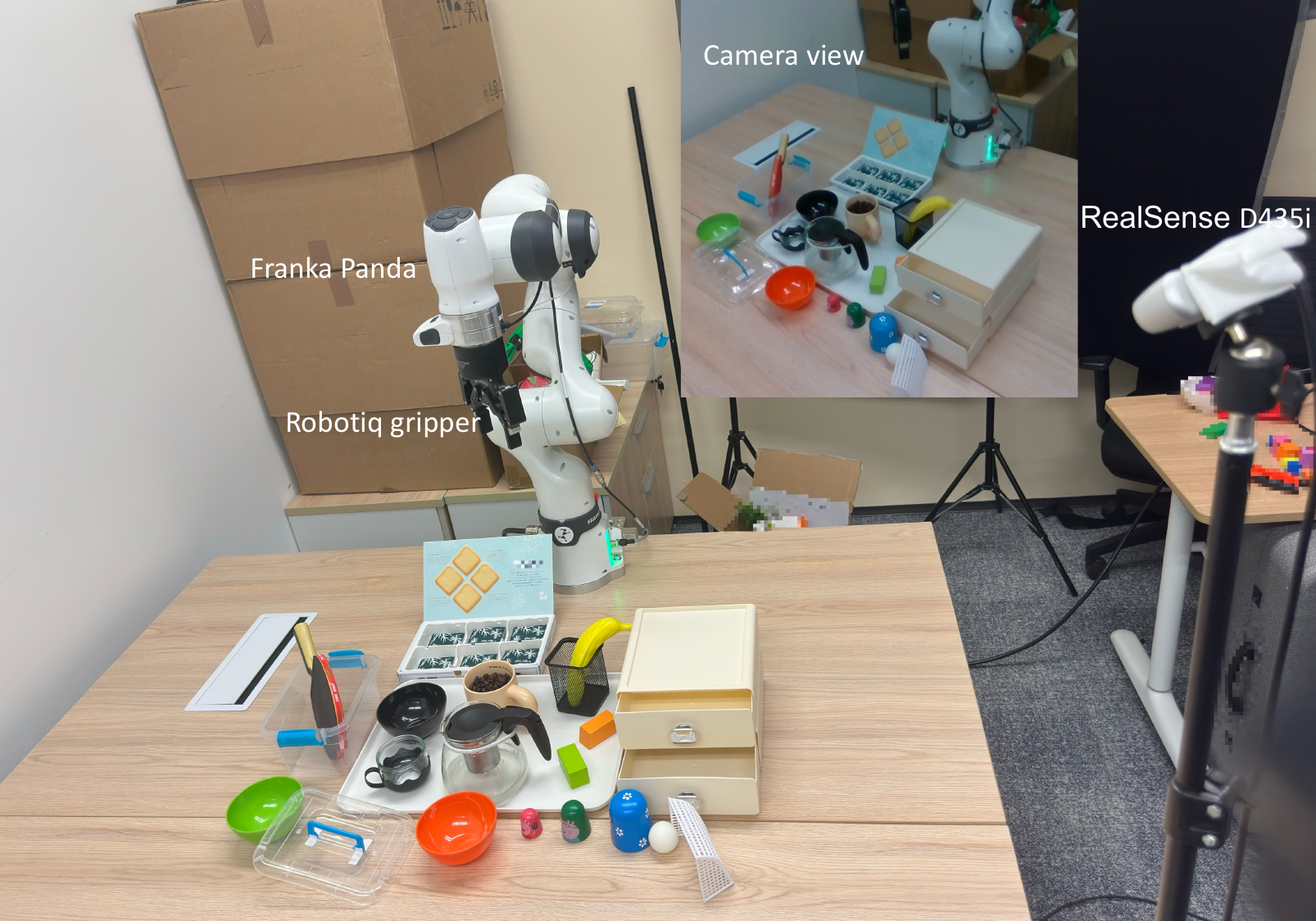}
    \caption{The experimental platform consists of Franka Emika Panda robot arm, Robotiq 2F-85 gripper and RealSense D435i positioned in third-person view.}
    \label{fig:robo_setup}
\end{figure}

\begin{figure*}[t]
    \centering
    \includegraphics[width=0.95\linewidth]{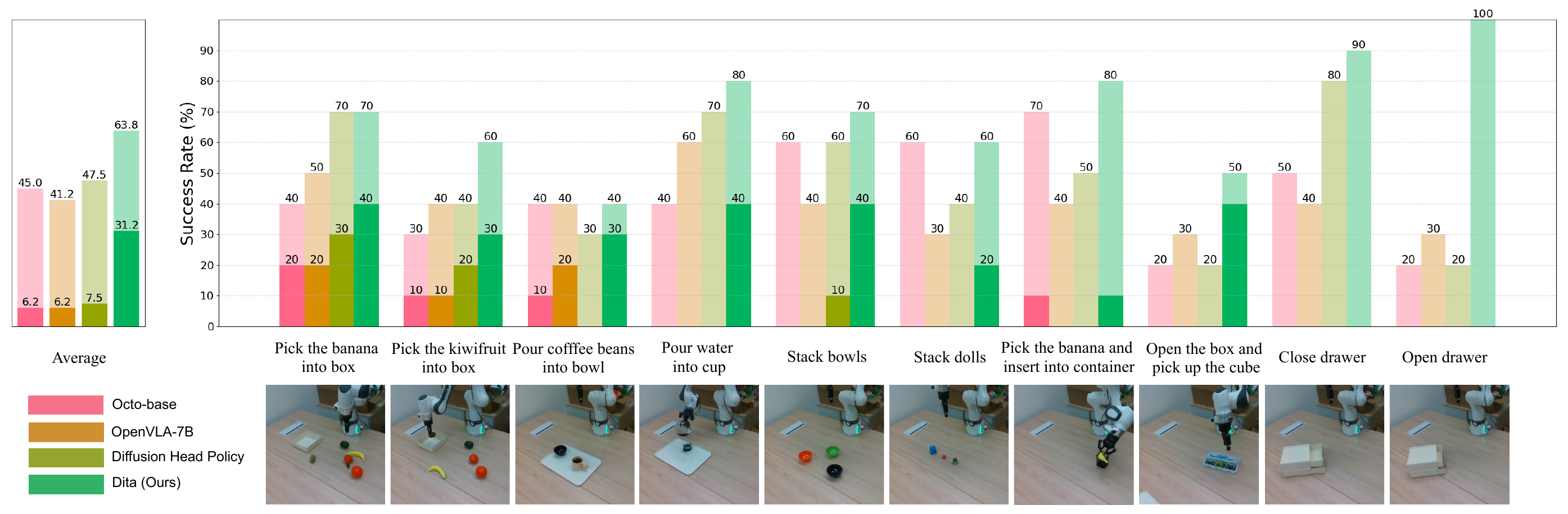}
    \caption{Quantitative results in real-robot experiments. Each task is manually divided into two sequential steps, except for the last two single-step tasks. In each stacked bar, the light-colored region represents the model's success rate in the first stage, while the dark-colored region indicates the contribution of second-stage success to the overall success rate. A larger proportion of the dark-colored region signifies a stronger capability of the model in completing long-horizon tasks. Since the open/close drawer tasks are single-step, they are excluded from the calculation of the average success rate. }
    \label{fig:franka_compare}
\end{figure*}

\subsection{Real-Robot Task Finetuning}

\noindent{\bf Setup.} To deploy the Dita model, as shown in Figure~\ref{fig:robo_setup}, the robot setup consists of a 7-DoF tabletop Franka Emika Panda robot arm and a Robotiq 2F-85 gripper. A RealSense D435i RGB-D camera, positioned approximately 1.5m away from the robot in a third-person view, captures RGB scenes with cluttered background at each inference timestep. Robot control is managed from a desktop computer running ROS, communicating with the model-deploy server with 1 NVIDIA A100 GPU. The system is operating under control frequency of 3Hz. Given the data domain gap between our robot platform and the pretrain dataset, we primarily evaluate Dita on 10-shot generalization for the following challenging tasks relevant to current VLA approaches:
\begin{itemize}
    \item {\bf Pick \& Place}. Two pick-and-place tasks with target object banana and kiwifruit are evaluated. 10 samples are collected for each task, with position variances introduced during evaluation to assess generalization performance.
    \item {\bf Pour}. We design two pouring tasks to evaluate the complex rotation finetuning: ``pour the coffee beans into the bowl'', and ``pour the water from the teapot into the cup''.
    \item {\bf Stack}. We design two stacking tasks for long-horizon pick-and-place: stacking three bowls and stacking three Russian dolls.
    \item {\bf Pick \& Rotation}. These skills combine two tasks: ``pick the banana and insert it into the small pen container'' and ``open the flip-top door box and then pick up the small cube''. 
    \item {\bf Pull \& Push}. We design the task ``open/close the drawer'' to evaluate pull and push abilities of our models.
    \item {\bf Long-Horizon Tasks.} We further devise several long-horizon tasks (more than 3 steps), including ``Pick up the bowl within the drawer and pour the coffee beans into the outside bowl'', ``pick up the racket and hit the ball into the goal'', ``open the top drawer, then pick the cube into the drawer, and finally close the drawer'', ``close the top drawer, then open the bottom drawer and put the bowl into the drawer, and finally close the drawer'' and ``open the box and move the green cube into the box then close the box'', to demonstrate the long-horizon manipulation ability of Dita (detailed in demo videos in Supplementary Materials).


\begin{figure}[t]
    \centering
    \includegraphics[width=0.99\linewidth]{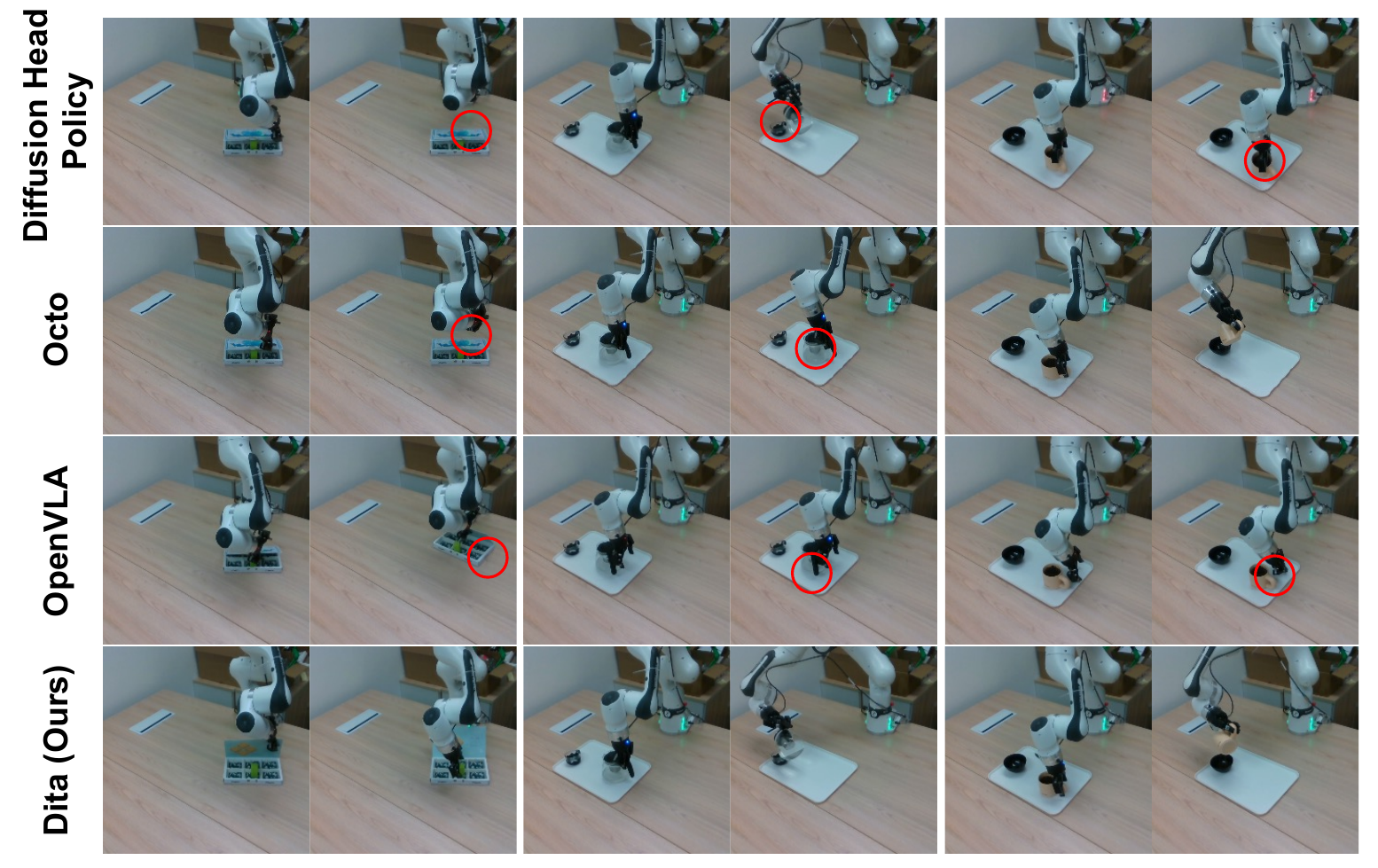}
    \caption{Qualitative comparison in real-robot experiments. Failures are highlighted with red circles. For a direct comparison, we initialize the layout consistently across all methods.}
    \label{fig:vis_comp_real}
    \vspace{-5mm}
\end{figure}



\end{itemize}

The diffusion policy~\cite{chi2023diffusion} has demonstrated strong capabilities in learning to mimic single tasks. Therefore, in addition to Octo and OpenVLA, we design a multimodal diffusion policy baseline based on a causal Transformer for comparison, which incorporates a similar diffusion head as described in Section~\ref{sec:sim_baseline}.




\noindent{\bf Optimization.} We finetune the Dita on the aforementioned multiple manipulation tasks, with data collected on the same platform, using LoRA~\cite{hu2021lora} for fair comparison and AdamW for 20,000 steps with image augmentations. The number of timesteps is set to 100 for DDPM~\cite{ho2020denoising}, and the batch size of 512.

\noindent{\bf 10-shot Finetuning.}  
We directly use the model pretrained on OXE datasets to evaluate 10-shot finetuning generalization for real-robot experiments. To ensure consistency with OpenVLA~\cite{kim2024openvla}, we finetune the network with one observation and one-step prediction. We compare the proposed method with OpenVLA~\cite{kim2024openvla} and Octo-base~\cite{team2024octo} models. Overall, Dita achieves a 63.8\% success rate on two-step tasks, with the second stage contributing nearly half, as shown in Figure~\ref{fig:franka_compare}. Dita consistently outperforms both Octo and OpenVLA, demonstrating superior performance on all complex tasks. For long-horizon tasks, OpenVLA effectively completes the first task but fails to handle the long-horizon task, such as completely misunderstanding the insert operation. In contrast, Octo performs better with rotation tasks and approaches the second step of the task more effectively. Supplementary materials provide extensive qualitative comparisons.

\noindent{\bf Variance Robustness.} 
To evaluate the robustness of Dita, we further validate its performance under different variances, including background changes, non-target object arrangements and lighting conditions. As illustrated in Figure~\ref{fig: teaser}, it is surprising to find that Dita not only excels in completing complex long-horizon tasks but also demonstrates resilience to a wide range of variations.


\subsection{Qualitative Comparison}

Figure~\ref{fig:vis_comp_real} presents a comparison between Dita, diffusion head baseline, Octo~\cite{team2024octo}, and OpenVLA~\cite{kim2024openvla} as evaluated in real-robot experiments under 10-shot finetuning setting. According to the visualized comparison, those baseline methods usually fail to grasp the correct position under the 10-shot setting, \eg, ``fail to insert the gap of the box'', ``grasp when the gripper is not in the correct grasping position (the hand of cup or teapot)''. Meanwhile, the baseline methods, \eg, the second raw (pouring water) in Figure~\ref{fig:vis_comp_real}, sometimes misunderstand the lifting action and get stuck after grasp. In contrast, Dita is able to effectively complete all the complex tasks with extreme 3D rotations. 


\section{Conclusion}


We present Dita, an architecture for generalist robot learning that leverages a Transformer-based diffusion model to denoise continuous action sequences through an in-context conditioning mechanism. By harnessing the scalability of Transformers, Dita effectively models diverse robot behaviors across extensive cross-embodiment datasets, enabling robust generalization across multiple simulation benchmarks within a unified framework. Additionally, Dita demonstrates strong few-shot adaptation capabilities, successfully transferring to novel real-world robot setups and long-horizon tasks with minimal in-domain samples. Notably, the model is clean, lightweight, and open-source, and its promising performance---achieved exclusively with a single third-person camera input---underscores its potential as a scalable and flexible solution for generalist policies.

\section{Acknowledgment}
This work is supported by the National Key R\&D Program of China (NO. 2022ZD0160201).



{
    \small
    \bibliographystyle{ieeenat_fullname}
    \bibliography{main}
}

\newpage
\appendix
\input{sec/X_suppl}

\end{document}

%% file: ICLR2025/math_commands.tex
\usepackage{amsmath,amsfonts,bm}

\def\eqref#1{equation~\ref{#1}}

\def\1{\bm{1}}

\def\va{{\bm{a}}}

\def\vx{{\bm{x}}}

\def\mI{{\bm{I}}}

\DeclareMathAlphabet{\mathsfit}{\encodingdefault}{\sfdefault}{m}{sl}
\SetMathAlphabet{\mathsfit}{bold}{\encodingdefault}{\sfdefault}{bx}{n}



%% file: sec/X_suppl.tex
\clearpage
\setcounter{page}{1}
\maketitlesupplementary

\section{Model and Training Scheme}
\label{sec:details}


The architecture of our model is illustrated in Figure 3. The language instruction is encoded using a pretrained CLIP model, with its encoder frozen during training. Input images are resized to $224 \times 224$ and processed by a pretrained DINOv2 model, with all parameters being finetuned. A Q-Former, trained from scratch with a depth of 4, is employed to reduce the dimensionality of the image features to a length of 32; within each block, text tokens are injected as FiLM conditions to augment the image features with linguistic information. The action is perturbed with noise via a DDPM scheduler with 100 timesteps, and a timestamp index is embedded using a sinusoidal positional embedding module. These multimodal inputs are then fed into a causal Transformer, which predicts the added noise. The Transformer adopts a LLaMA2-style architecture, trained from scratch, comprising 12 self-attention blocks with a hidden size of 768. All components are trained except for the CLIP text encoder. In total, the model comprises 334M parameters, with 221M being trainable. Achieving this level of performance with such a compact model represents a pioneering advancement in the field, underscoring the efficacy of the architectural design.

\section{Simulation Benchmarks}

\subsection{SimplerEnv} \label{sec:simple_supp}

\begin{figure}
    \centering
    \includegraphics[width=1.\linewidth]{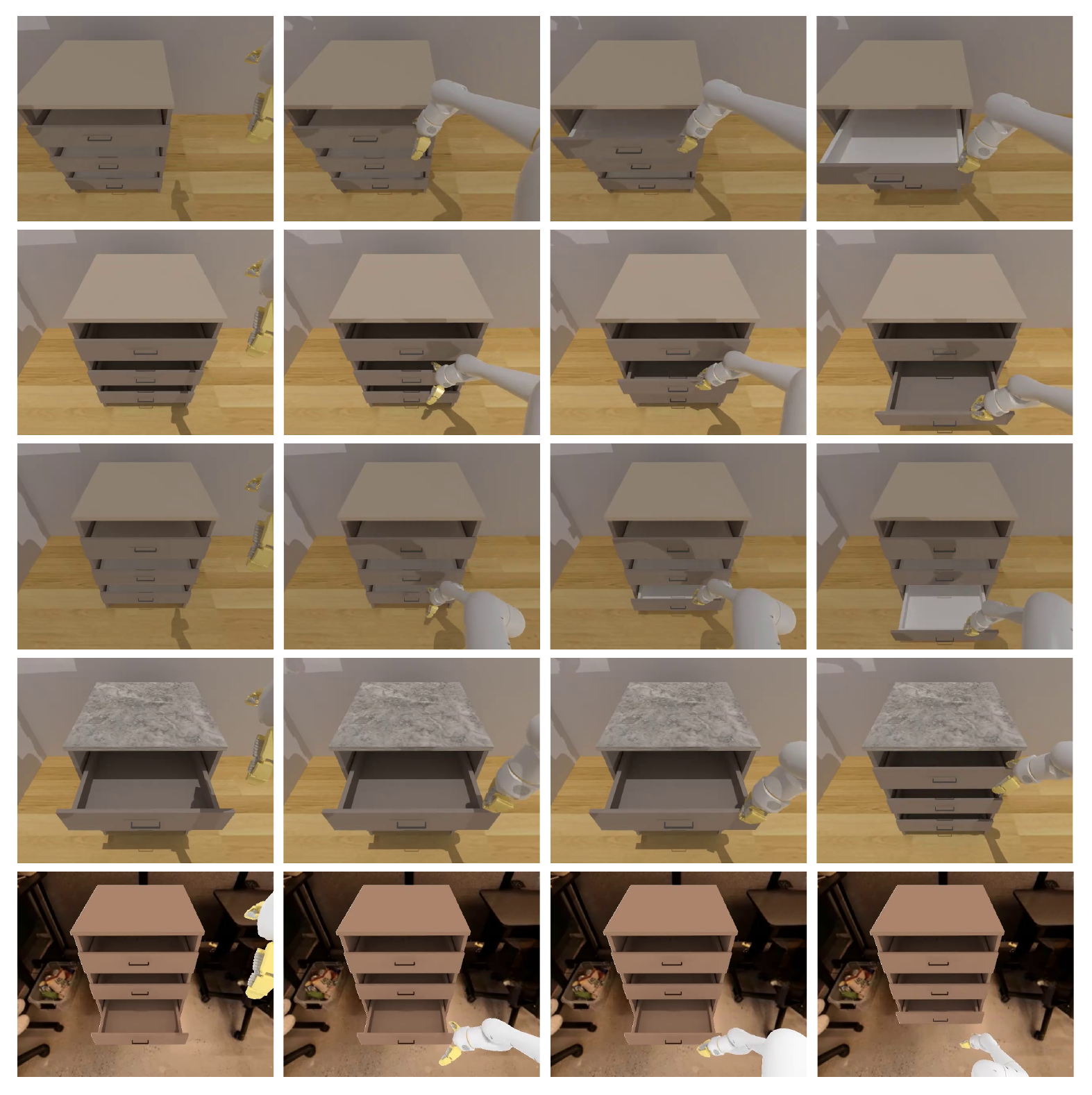}
    \caption{Qualitative results of Dita under variances in Google Robot.}
    \label{fig:vis_comp_googlerobot}
\end{figure}

\noindent{\bf Results.} As described in Figure~\ref{fig:vis_comp_googlerobot}, leveraging the in-context conditioning design, Dita exhibits enhanced robustness, relying solely on third-person view images to detect subtle nuances and generate more reliable actions.

\subsection{LIBERO}

LIBERO comprises four subtasks: LIBERO-SPATIAL, LIBERO-OBJECT, LIBERO-GOAL, and LIBERO-100, each designed to evaluate different model capabilities. LIBERO-SPATIAL assesses spatial relationship understanding, containing data with identical object sets but varying layouts. LIBERO-OBJECT evaluates object transferability, featuring data with consistent layouts but different object sets. LIBERO-GOAL examines task comprehension and transferability, maintaining the same object sets and layouts while varying tasks. LIBERO-100 is further divided into LIBERO-90 and LIBERO-10 (also referred to as LIBERO-LONG), designated for policy pretraining and long-horizon task evaluation, respectively. LIBERO-100 encompasses a diverse range of objects, layouts, and backgrounds, providing a comprehensive benchmark for generalization in robot learning.


\begin{figure*}[t]
    \centering
    \includegraphics[width=1.\linewidth]{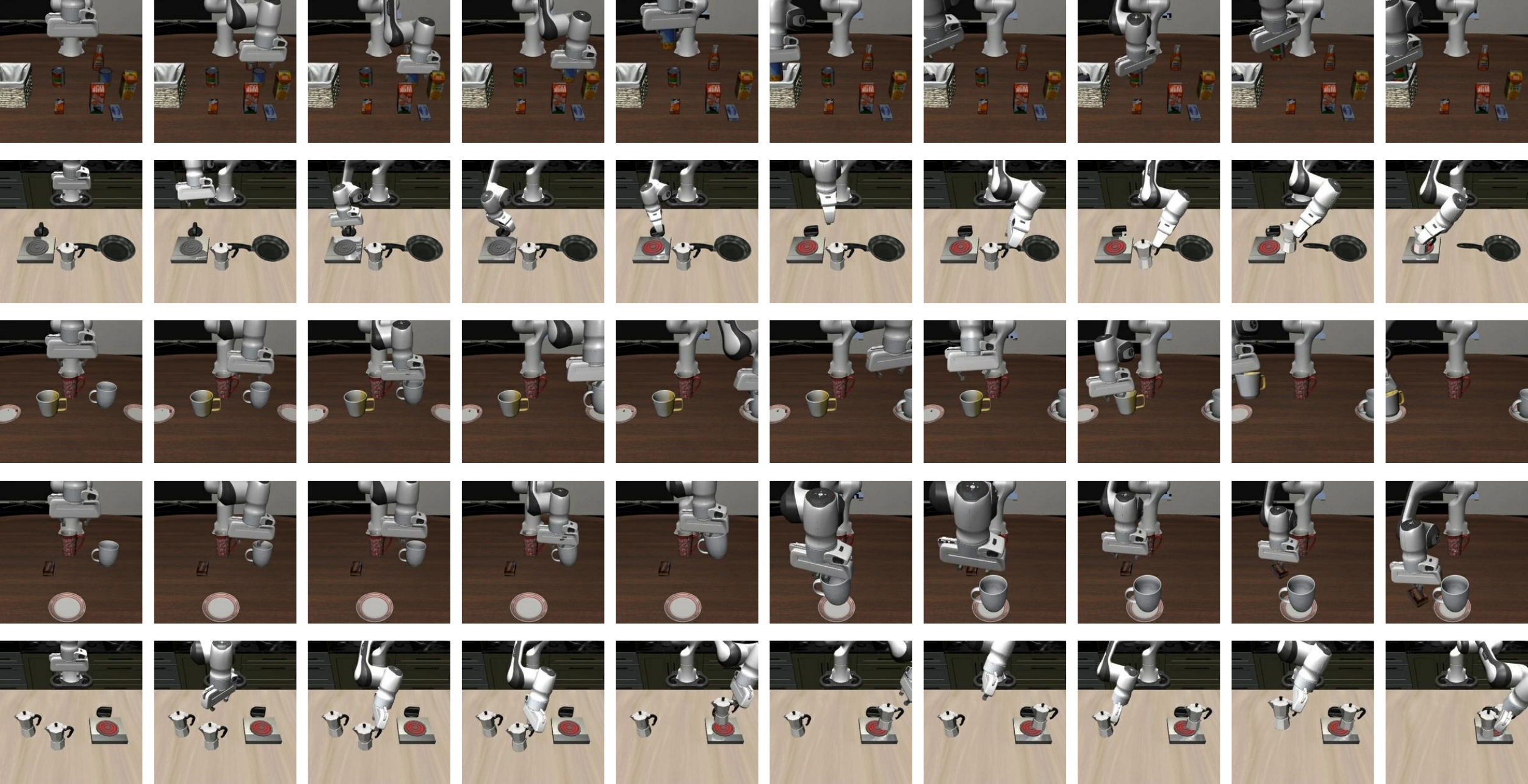}
    \caption{Qualitative results of Dita on LIBERO benchmark.}
    \label{fig:vis_comp_libero}
\end{figure*}

\noindent{\bf Optimization.} We optimize the network using AdamW for 100,000 steps on LIBERO. The learning rate is set to $1e-4$ for LIBERO-SPATIAL, LIBERO-OBJECT, and LIBERO-GOAL, and $5e-4$ for LIBERO-LONG. Across all sub-datasets, a half-cycle cosine scheduler is applied to decay the learning rate. Denoising timestamps are set to 100 during finetuning, and training is conducted with a batch size of 512 across 8 NVIDIA A100 GPUs.

\noindent{\bf Results.} 
Table~\ref{tab:libero_supp} shows that Dita achieves a success rate of 77.93\% on the most challenging task in LIBERO, \ie, SPATIAL-LONG. We argue that the Droid dataset~\cite{khazatsky2024droid} serves as a more suitable pretraining dataset for LIBERO, as our model (334M) lacks the capacity to fully accommodate the entire OXE dataset. We anticipate that performance on the OXE-pretrained model can be significantly improved by scaling up the model size.

\begin{table}[t]
\setlength\tabcolsep{4pt}
\caption{Comparison with Diffusion Policy~\cite{chi2023diffusion}, Octo~\cite{team2024octo}, and OpenVLA~\cite{kim2024openvla} on LIBERO~\cite{liu2024libero}. Dita (OXE) denotes the use of a pretrained model on OXE, while Dita (Droid) refers to the use of a pretrained model on Droid. 
}
\label{tab:libero_supp}
\small
\begin{center}
\begin{tabular}{l|c}
\toprule
{\bf Method}  & {\bf LIBERO-LONG} \\
\hline 
Diffusion Policy*~\cite{chi2023diffusion} &  50.5\%  \\
Octo~\cite{team2024octo} & 51.1\%  \\
OpenVLA~\cite{kim2024openvla} & 53.7\% \\
\hline
Dita (pretrained on OXE) & 63.8\% \\
Dita (pretrained on Droid) & \bf 77.9\% \\

\bottomrule
\end{tabular}
\end{center}

\end{table}

\subsection{CALVIN}

\noindent{\bf Setup.} 
We directly apply the proposed method to CALVIN using a single static RGB camera to predict the end-effector action, which includes three dimensions for translation, three dimensions for Euler angle rotation, and one dimension for gripper position (open or close). We evaluate Dita and $\mathcal{E}_{S}^{Diff}$ on CALVIN, leveraging the pretrained model on the OXE dataset to initialize the model for CALVIN.

\begin{figure}
    \centering
    \includegraphics[width=1.\linewidth]{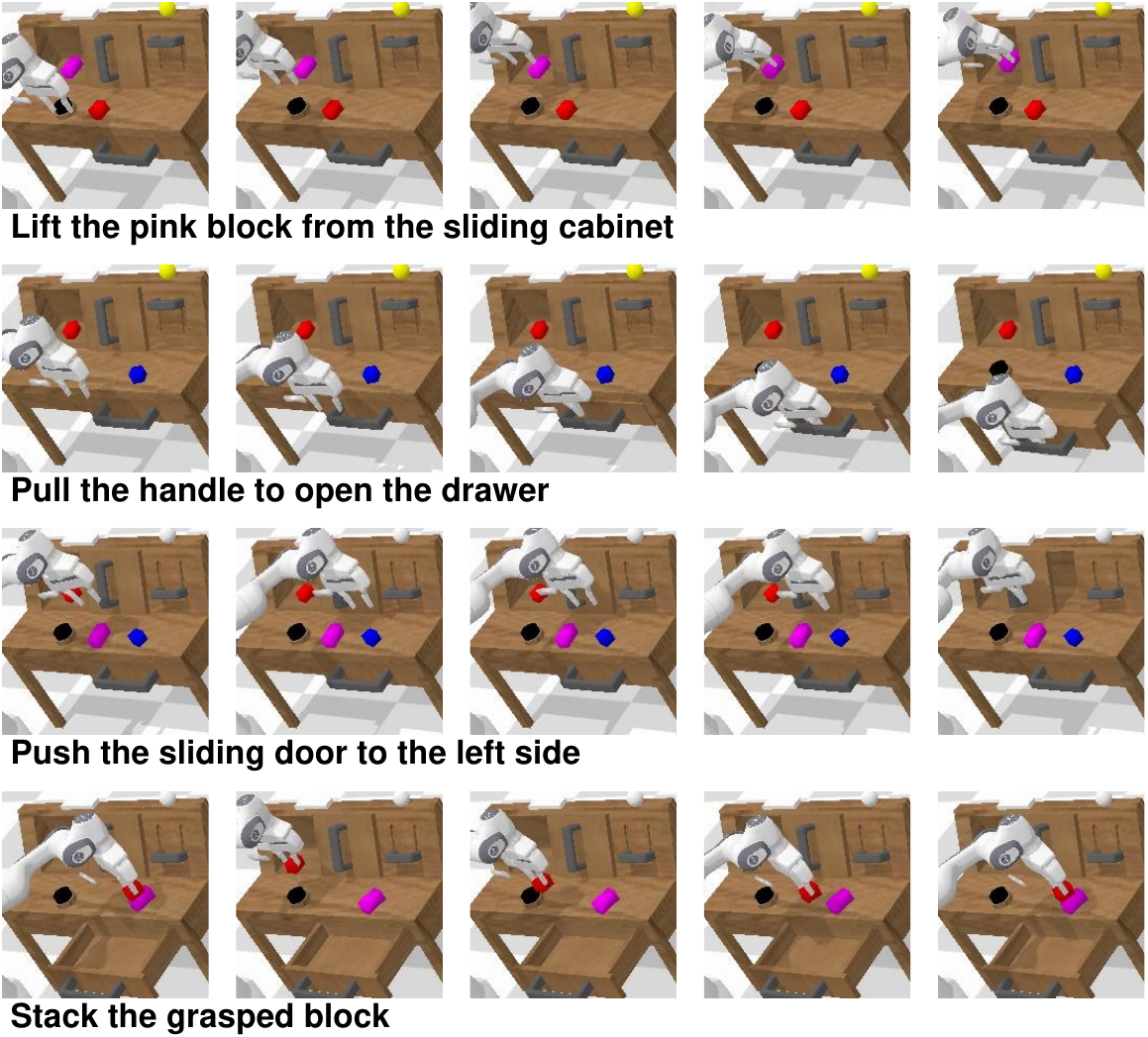}
    \caption{Qualitative results of Dita on CALVIN ABC$\rightarrow$D benchmark.}
    \label{fig:vis_comp_calvin}
\end{figure}

\noindent{\bf Optimization.} For each training iteration, the model predicts 10 future action chunks supervised by MSE loss. An AdamW optimizer is used together with a decayed learning rate with half-cycle cosine scheduler after several steps of warming up. The learning rate is initialized as $1e-4$. Model is trained for 15 epochs with batch size of 128 across 4 NVIDIA A100 GPUs. 

\subsection{Maniskill2}

ManiSkill2~\cite{gu2023ManiSkill2}, the next generation of the SAPIEN ManiSkill benchmark~\cite{mu2021maniskill}, serves as a widely recognized platform for assessing the generalized manipulation capabilities of embodied models. It encompasses 20 distinct manipulation task families and over 4M demonstration frames across various configurations. Leveraging ManiSkill2, we establish a \textit{novel camera view generalization} benchmark to evaluate the effectivenes of Dita.

{\bf Setup.} To construct the benchmark, we select 5 tasks (PickCube-v0, StackCube-v0, PickSingleYCB-v0,PickClutterYCB-v0, PickSingleEGAD-v0) from ManiSkill2 and create a camera pool comprising 300K random cameras. 20 cameras are sampled each time to render each trajectory, resulting in over 40K trajectories, which are utilized to train Dita from scratch. The generated dataset is divided into training and validation sets with a 19:1 ratio, ensuring that each category in task family, and trajectories rendered from different camera views are assigned to both, thereby preventing data leakage. During training, the number of data samples is balanced across task families by duplicating trajectories for task families with fewer samples. To construct a closed-loop evaluation dataset, we randomly sample 100 trajectories from the validation set for each task family. This evaluation dataset with 500 trajectories is used to assess the success rate for each task family and demonstrate the camera-view generalization capabilities of Dita.

\noindent{\bf Optimization.} The network is optimized using AdamW for 50,000 steps on ManiSkill2, with a learning rate set to $1e-4$. The number of denoising timestamps is set to 100, and the batch size is 1024 distributed across 16 NVIDIA A100 GPUs.

\begin{figure}
    \centering
    \includegraphics[width=1.\linewidth]{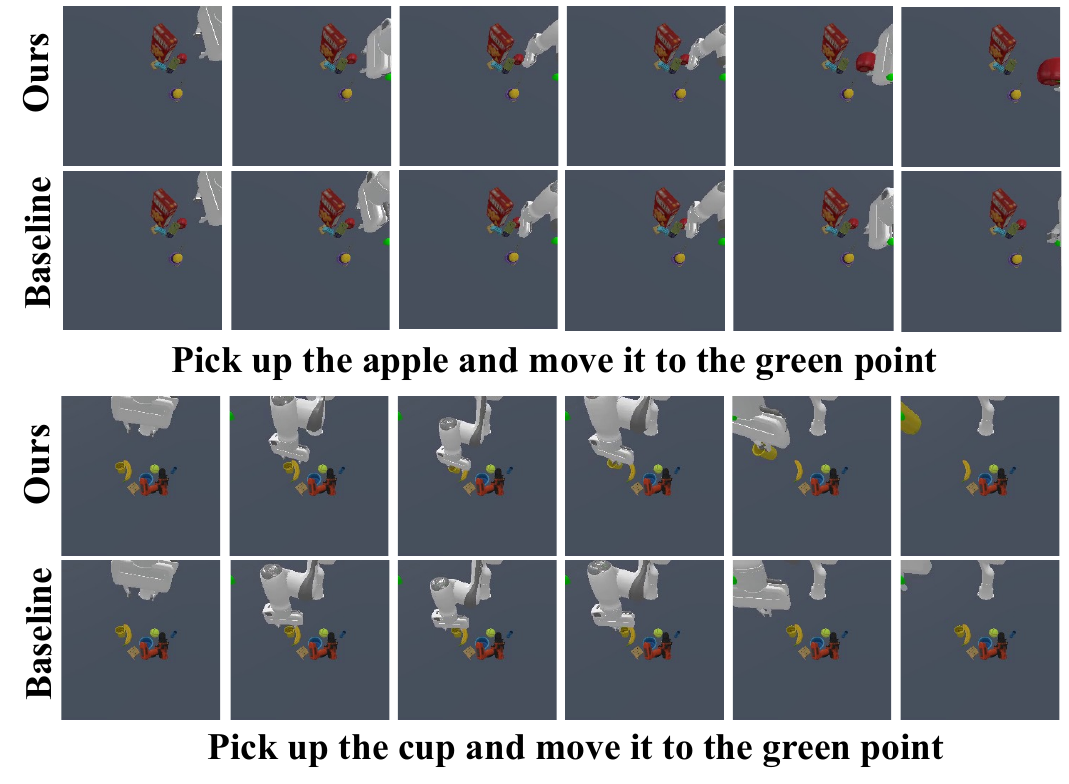}
    \caption{Qualitative comparison between Dita (top) and Diffusion Action Head baseline $\mathcal{E}_{\theta \sim s}^{Diff}$ (bottom) on ManiSkill2 (PickClutterYCB).}
    \label{fig:vis_comp_maniskill2}
\end{figure}

\section{Real-Robot Experiments}
\subsection{Real-Robot Setup}

\noindent{\bf Optimization.} We apply image augmentations using ColorJitter from the torchvision library, with brightness set to 0.3, contrast ranging from 0.7 to 1.3, saturation ranging from 0.7 to 1.3, and hue set to 0.07. Further details are provided in the code.

\noindent{\bf Variance Robustness.} To evaluate the robustness of Dita, we further validate its performance under different variances, including:
\begin{itemize}
    \item \textit{Background changes.} The background includes both the tabletop color and the backdrop. We introduce variance in both aspects by using tablecloths in colors different from the tabletop and a black backdrop.
    \item \textit{Non-target object arrangements.} We randomly place non-target objects in arbitrary poses within the robot's workspace to create a cluttered scene, whereas it remains clean during demonstration recording.
    \item \textit{Lighting conditions.} We modify the lighting by turning off one of the two lights in the room to introduce variation in illumination.
\end{itemize}


\subsection{Details of Real-Robot Tasks}
\label{sec:tasks}

In addition to the fundamental tasks used for quantitative comparison with prior approaches, we incorporate complex long-horizon tasks that previous methods fail to complete for illustrative purposes. Below, we present all tasks along with their step-wise decomposition.

\begin{itemize}

 \item \textit{ Pick the banana into the box.} We divided this task into two steps: first, successfully picking up the banana, and second, successfully placing it into the box.

 \item \textit{ Pick the kiwifruit in the box.} We divided this task into two steps: first, successfully picking up the kiwifruit, and second, successfully placing it into the box.

 \item \textit{ Pouring the coffee beans into the bowl.} This task is divided into two steps: first, successfully picking up the cup, and second, successfully pouring the coffee beans within the cup into the box.

 \item \textit{ Pouring the water from the teapot into the cup.} This task is divided into two steps: first, successfully picking up the teapot, and second, successfully pouring the water into the cup.

 \item \textit{ Stacking three bowls.} This task is divided into two steps: Stacking the first bowl successfully, and second stacking the left bowl into previous stacked bowls.

 \item \textit{ Stacking three nesting dolls.} This task is divided into two steps: Stacking the first two small dolls successfully, and second stacking the large doll into previous stacked dolls.

 \item \textit{ Pick the banana and insert into the small pen container.} This task is divided into two steps: first, successfully picking up the banana, and second, successfully inserting the banana into the pen container.

 \item \textit{ Open the Flip-top door box and the pick up the small cube inside.} This task is divided into two steps: first, successfully open the door box, and second, successfully picking up the small cube.

 \item \textit{ Open the drawer.} This task has only one step.

 \item \textit{ Close the drawer.} This task has only one step.

 \item \textit{ Pick up the bowl within the drawer and pouring the coffee beans into the outside bowl.} This is a long horizon task, and we demonstrate it with video in the supplementary appendix given that previous approaches fail to complete the task.

 \item \textit{ Pick up the racket and hit the ball into the goal} This is a long horizon task, and we demonstrate it with video in the supplementary appendix given that previous approaches fail to complete the task.

 \item \textit{ Open the top drawer, then pick the cube into the drawer, and finally close the drawer.} This is a long horizon task, and we demonstrate it with video in the supplementary appendix given that previous approaches fail to complete the task.

 \item \textit{ Close the top drawer, then open the bottom drawer and put the bowl into the drawer, and finally close the drawer.} This is a long horizon task, and we demonstrate it with video in the supplementary appendix given that previous approaches fail to complete the task.

 \item \textit{ Open the box and move the green cube into the box then close the box.} This is a long horizon task, and we demonstrate it with video in the supplementary appendix given that previous approaches fail to complete the task.
 
\end{itemize}

\subsection{Finetuing Details}


We adhere to~\cite{kim2024openvla} and employ LoRA for 10-shot finetuning. However, Dita comprises only 221M trainable parameters, with merely 5\% (approximately 11M) remaining trainable under LoRA finetuning. We contend that this limited capacity is inadequate to effectively accommodate image augmentations, thereby compromising robustness against environmental variances. To this end, we evaluate robustness through full finetuning and observe a substantial improvement in the success rate for long-horizon tasks, alongside greater resilience to variances such as background changes, non-target object arrangements, and lighting conditions. Quantitatively, full finetuning achieves a success rate of 20\%, whereas LoRA finetuning fails to complete tasks under extreme variances.

\section{Analysis, Ablations, and Discussions}
\label{sec:analysis}

\subsection{Practices on reproducing Octo and OpenVLA }
\label{sec:analysis:long}



We observe that OpenVLA~\cite{kim2024openvla} demonstrates superior pick-up performance compared to Octo~\cite{team2024octo}. However, for tasks requiring the learning of rotational operations, such as opening a box, Octo achieves better performance. We attribute this to Octo’s ability to predict continuous actions, which are less sensitive to action normalization, whereas OpenVLA relies on action discretization based on action statistics. We compute the statistics from the 10-shot training samples across all tasks and find it challenging to obtain suitable statistics for discretization values, which are unnecessary for the diffusion policy.

\begin{figure}
    \centering
    \includegraphics[width=0.9\linewidth]{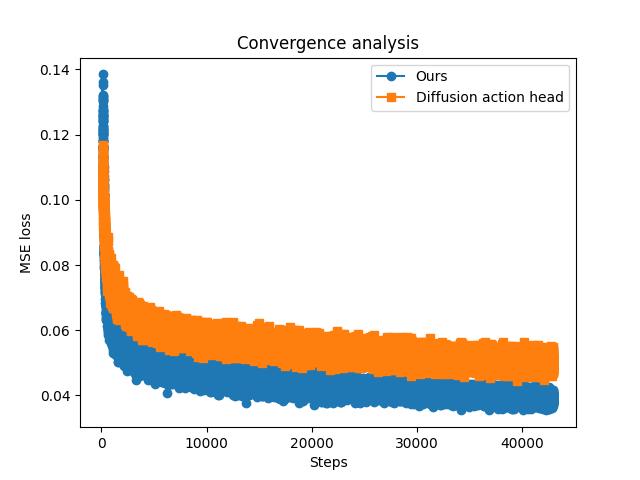}
    \caption{Convergence Analysis on OXE dataset~\cite{brohan2023rt}. The blue line is DiT Policy, and the orange line is Diffusion action head strategy with the same number of parameters.}
    \label{fig:convergence}
\end{figure}

\begin{table}[t]
\setlength\tabcolsep{1pt}
 \centering
 \caption{Additional experiments on Calvin (ABC$\rightarrow$D). `aug' indicates image augmentation during finetuning.}
  \label{tab:dit_ab}
  
  \small
  \begin{tabular}{c|ccccc|c}
    \toprule    
    \multirow{2}{*}{\bf Method} & \multicolumn{6}{|c}{\bf No. Instructions in a Row (1000 chains)} \\
    \cline{2-7}
     & 1 & 2 & 3 & 4 & 5 & Avg.Len. \\ \hline


Dita w/o aug & 94.5\%  &  82.5\%  & 72.8\% & 61.3\%  & 50.0\% & 3.61 \\

Dita w/aug & 92.3\% & 83.8\% & 75.8\% & 67.3\% & 69.0\% & {\bf 3.78} \\
\hline

\multicolumn{7}{c}{w/o pretraining w/ aug}  \\ \hline
Dita (2 layers) & 75.8\% & 47.3\% & 29.3\% & 19.0\% & 12.5\% & 1.84 \\
Dita (6 layers) & 80.5\% & 60.0\% & 39.0\% & 26.3\%  &15.0\% & 2.21 \\

Dita (12 layers)  & 84.5\% & 65.5\% & 48.8\% & 34.5\% & 22.5\% & 2.56\\
Dita (24 layers) & 87.8\% & 67.5\% & 48.5 \%  & 34.5 \%  & 23.0\% & 2.61 \\

\hline

Dita  & 84.5\% & 65.5\% & 48.8\% & 34.5\% & 22.5\% & 2.56\\
AdaLN DiT [17]  & 68.3\% & 40.0\% & 21.0\% & 11.3\% & 5\% & 1.45 \\


\bottomrule
\end{tabular}
\end{table}

\subsection{Convergence Analysis}
\label{sec:analysis:converge}

Figure~\ref{fig:convergence} illustrates the convergence comparison between the diffusion head baseline $\mathcal{E}_{\theta}^{Diff}$ and Dita. Dita achieves clearly faster convergence than $\mathcal{E}_{\theta}^{Diff}$. We believe this further highlights the scalability of Dita.

\subsection{Failure Analysis}

\begin{figure}
    \centering
    \includegraphics[width=0.9\linewidth]{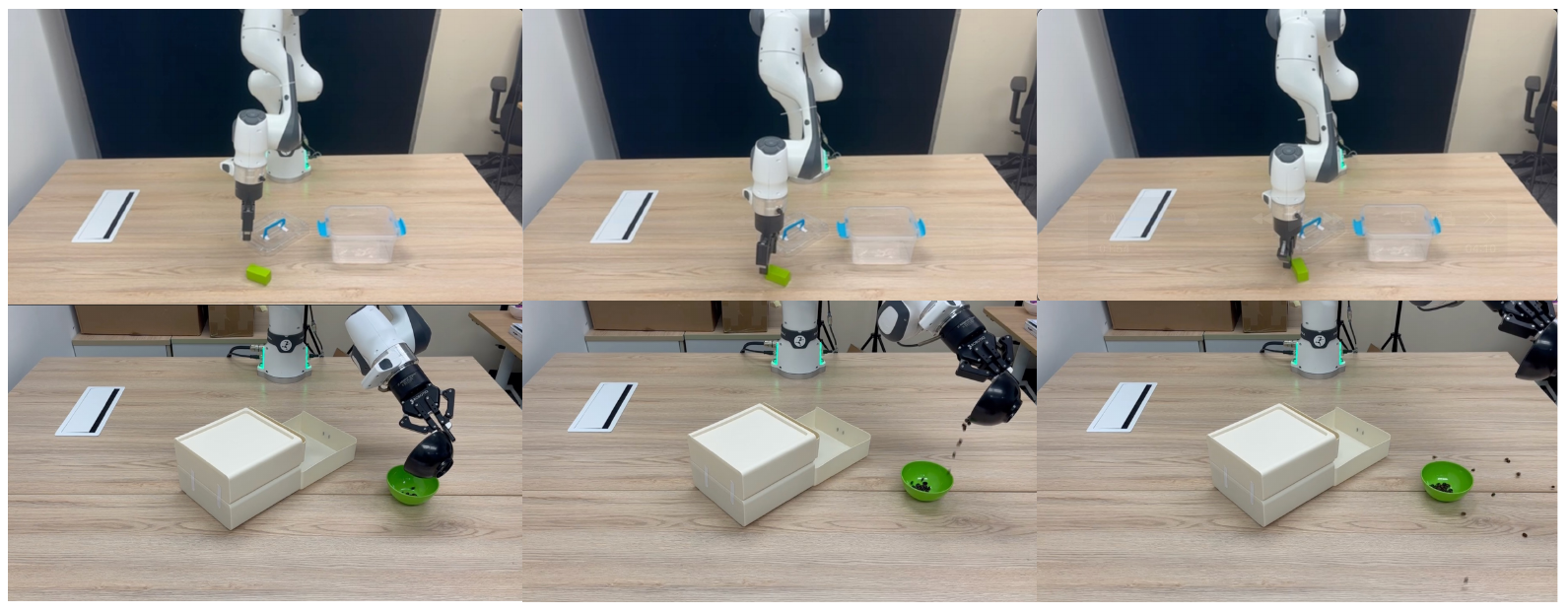}
    \caption{Failure Cases Analysis.}
    \label{fig:failure}
\end{figure}

Figure~\ref{fig:failure} illustrates two representative failure cases. The first case involves a failed grasp during execution. Although the model is capable of recovering and retrying, the failure suggests an inability to approach the correct position for a successful pickup. The second case concerns pouring while in motion, where the model tends to spill the contents due to unstable handling.

%


%


 \begin{table}[t]
 \setlength\tabcolsep{4pt}
 \caption{The ablation study on the learning rate scheduler in the Calvin benchmark.}

 \centering
  \label{tab:calvin_lr}
  \small
  
  \begin{tabular}{c|c|c|c|c|c|c}
   \toprule
     {\bf Strategy } & \multicolumn{6}{|c}{\bf No. Instructions in a Row (1000 chains)} \\ 
    \hline
     w lr decay & {\bf 94.5}\%  & {\bf 82.5}\%  & {\bf 72.8}\% & {\bf 61.3}\% & {\bf 50.0}\% & {\bf 3.61} \\
     w/o lr decay & 91.8\%  & 80.0\%  & 68.0\% & 56.9\% & 45.9\%  & 3.43 \\
   \bottomrule
  \end{tabular}

\end{table}

\subsection{More Ablations}
\label{sec:diffusion_baselines}

\noindent{\bf Learning Rate Scheduler.}
As outlined in the main text, we utilize a standard learning rate scheduler to decay the learning rate during the experiments in Calvin, rather than using a fixed learning rate of $1e-4$ as in the pretraining stage. This adjustment results in a slight performance improvement, as shown in Table~\ref{tab:calvin_lr}.


\begin{table}[t]
\setlength\tabcolsep{2pt}
\caption{More action designs w/o pretraining on Calvin (ABC$\rightarrow$D). MDT is from issue 9 of its GitHub repo and GR-MG.
}
\vspace{-3mm}

\centering
\label{tab:calvin_baselines}
\small

\begin{tabular}{c|c|c|c|c|c|c}
\toprule
 {\bf Methods } & \multicolumn{6}{|c}{No. Instructions in a Row (1000 chains)} \\ 
\hline
MDT*~\cite{reuss2024multimodal} & 61.7\% & 40.6\% & 23.8\% & 14.7\% & 8.7\% & 1.54 \\
\hline 
Unet1D head~\cite{chi2023diffusion} & 76.8\% & 46.5\% & 28.8\% & 18.5\% & 10.0\% & 1.80 \\

Transformer head~\cite{chi2023diffusion} & 75.8\% &  44.8\% & 26.5\% &16.5\% & 8.0\% & 1.72 \\
%

8-layer MLP head & 69.8\% & 42.5\% & 26.3\% & 16.8\% & 11.0\% & 1.66\\

3-layer MLP head & 75.5\%  & 44.8\%  & 25.0\% & 15.0\% & 7.5\% & 1.68 \\

\hline

Single token act chunks & 56.5\% & 18.3\% & 6.0\% & 2.8\% & 0.8\% & 0.84 \\

%
Ours & {\bf 89.5\%}  & {\bf 63.3\%}  & {\bf 39.8\%} & {\bf 27.3\%} & {\bf 18.5\%} & {\bf 2.38} \\

\bottomrule  
\end{tabular}
\end{table}

\begin{table}[t]
\caption{The effect of the number of execution steps (\# Steps) on ManiSkill2.}
\label{tab:maniskill_ab1}
\small
\centering
\begin{tabular}{c|c|c|c|c|c}
\toprule
\bf{\# Steps} & 1 &  2  & 4  & 8 & 16 \\
\hline
All & {\bf 61.6}\% & 60.8 \% & 60.6 \% & 60.0 \% & 58.0 \%  \\
\bottomrule
\end{tabular}
\end{table}

\begin{table}[h]
\setlength\tabcolsep{1.5pt}
\caption{Ablation study of shuffle buffer size on SimplerEnv (both math and variant results of Google Robot~\cite{brohan2022rt}).}
\label{tab:ab_buffle}

\small
\begin{center}
\begin{tabular}{c|cc}
\toprule
\multirow{2}{*}{\bf Shuffle Buffer Size}  &\multicolumn{2}{c}{\bf coke$\_$can} \\
 & { match } & {variant} \\
\hline 

128000 & 71.2\% & 73.6\% \\
256000 & \bf 83.7\% & {\bf 85.5\%}\\

\bottomrule
\end{tabular}

\end{center}
\end{table}

\noindent{\bf Diffusion Head.} We implement several policies inspired by the core idea of the diffusion action head, which differ slightly from Octo~\cite{team2024octo} in predicting action chunks. Specifically, Octo~\cite{team2024octo} flattens action chunks into a single vector with a unified embedding. For instance, when predicting 8 actions, it generates a $8 \times 7 = 56$-dimensional vector. In contrast to the Octo-style diffusion action head, we adopt a diffusion action head, akin to Diffusion Loss~\cite{li2024autoregressive} and Diffusion Force~\cite{chen2024diffusion}, which are more effective. We evaluate multiple diffusion heads, including Unet1D, Transformer, and MLP on Calvin without pretraining.



Table~\ref{tab:calvin_baselines} shows that Dita achieves the better generalization on Calvin (ABC$\rightarrow$D), compared to other diffusion head strategies~\cite{bharadhwaj2024roboagent,zhao2023learning,chi2023diffusion}.

\noindent{\bf Execution steps.}
Since Dita can anticipate multiple future actions, we can execute multiple steps within a single inference. Here, we analyze the impact of execution steps under a model with a trajectory length of 32, as presented in Table~\ref{tab:maniskill_ab1}. The ablation study reveals that shorter execution steps yield slightly better results than longer ones; that is, the further the prediction extends from the current frame, the lower its accuracy. Nevertheless, the slight performance drop demonstrates that even with only 2-frame image observations, Dita can generate reliable action trajectories, underscoring its scalability.

\noindent{\bf Shuffle Buffer Size.} 
The shuffle buffer size of TensorFlow datasets has a significant impact on performance. Following OpenVLA~\cite{kim2024openvla, padalkar2023open}, we utilize TensorFlow datasets for network optimization, where the shuffle buffer size similarly influences performance (Table~\ref{tab:ab_buffle}), as observed in Octo~\cite{team2024octo}.

\noindent{\bf Denoising steps.} Typically, diffusion models require multiple denoising steps in image generation~\cite{rombach2021stablediffusion}. For diffusion-based policies in robot learning, the number of denoising steps during inference can impact control frequency. Surprisingly, we find that DDIM significantly reduces the denoising steps to 10 without compromising performance on the "Pick Coke" task as described in Table~\ref{tab:denosing_step}. With only 2 denoising steps, the model still achieves a 70.4\% success rate. We attribute model performs best with 10 steps to the reduction of overfitting when fewer denoising steps are used. Unlike image generation, the action dimension in robot learning is much smaller, allowing effective denoising with fewer steps without requiring advanced techniques~\cite{song2023consistency}. These results suggest that the in-context conditioning used by Dita does not hinder inference speed.

\begin{table}[t]
\setlength\tabcolsep{4pt}
\caption{The impact of the number of denoising steps (\# Steps) of DDIM on Google Robot Simulation during inference, trained with 1000 DDPM denoising steps.}
\label{tab:denosing_step}
\small
\begin{center}
\begin{tabular}{l|c|c|c|c|c|c}
\toprule
{\bf \# Steps} &  {\bf 100} & {\bf 50} & {\bf 20} & {\bf 10} & {\bf 5} & {\bf 2} \\
\hline
Pick Coke (variant) & 76.4 & 79.1 & {\bf 85.5} &  85.3 & 82.7 & 70.4 \\
Pick Coke (match) & 79.7 & 83.3 & {\bf 83.7}  & 82.0 & 82. & 73.3 \\
Move Near (variant) & 52.1 & 66.0 & {\bf 73.0} & 69.5 & 63.5 & 51.6\\
Move Near (match) & 49.1 & 72.0 & {\bf 76.0} & 74.0 & 72.0 & 65.0\\
\bottomrule
\end{tabular}
\end{center}
\end{table}

%% file: main.bbl
\begin{thebibliography}{85}
\providecommand{\natexlab}[1]{#1}
\providecommand{\url}[1]{\texttt{#1}}
\expandafter\ifx\csname urlstyle\endcsname\relax
  \providecommand{\doi}[1]{doi: #1}\else
  \providecommand{\doi}{doi: \begingroup \urlstyle{rm}\Url}\fi

\bibitem[Belkhale et~al.(2024)Belkhale, Ding, Xiao, Sermanet, Vuong, Tompson, Chebotar, Dwibedi, and Sadigh]{belkhale2024rt}
Suneel Belkhale, Tianli Ding, Ted Xiao, Pierre Sermanet, Quon Vuong, Jonathan Tompson, Yevgen Chebotar, Debidatta Dwibedi, and Dorsa Sadigh.
\newblock Rt-h: Action hierarchies using language.
\newblock \emph{arXiv preprint arXiv:2403.01823}, 2024.

\bibitem[Beyer et~al.(2024)Beyer, Steiner, Pinto, Kolesnikov, Wang, Salz, Neumann, Alabdulmohsin, Tschannen, Bugliarello, et~al.]{beyer2024paligemma}
Lucas Beyer, Andreas Steiner, Andr{\'e}~Susano Pinto, Alexander Kolesnikov, Xiao Wang, Daniel Salz, Maxim Neumann, Ibrahim Alabdulmohsin, Michael Tschannen, Emanuele Bugliarello, et~al.
\newblock Paligemma: A versatile 3b vlm for transfer.
\newblock \emph{arXiv preprint arXiv:2407.07726}, 2024.

\bibitem[Bharadhwaj et~al.(2024)Bharadhwaj, Vakil, Sharma, Gupta, Tulsiani, and Kumar]{bharadhwaj2024roboagent}
Homanga Bharadhwaj, Jay Vakil, Mohit Sharma, Abhinav Gupta, Shubham Tulsiani, and Vikash Kumar.
\newblock Roboagent: Generalization and efficiency in robot manipulation via semantic augmentations and action chunking.
\newblock In \emph{2024 IEEE International Conference on Robotics and Automation (ICRA)}, pages 4788--4795. IEEE, 2024.

\bibitem[Black et~al.(2023)Black, Nakamoto, Atreya, Walke, Finn, Kumar, and Levine]{black2023zero}
Kevin Black, Mitsuhiko Nakamoto, Pranav Atreya, Homer Walke, Chelsea Finn, Aviral Kumar, and Sergey Levine.
\newblock Zero-shot robotic manipulation with pretrained image-editing diffusion models.
\newblock \emph{arXiv preprint arXiv:2310.10639}, 2023.

\bibitem[Black et~al.(2024)Black, Brown, Driess, Esmail, Equi, Finn, Fusai, Groom, Hausman, Ichter, et~al.]{black2024pi_0}
Kevin Black, Noah Brown, Danny Driess, Adnan Esmail, Michael Equi, Chelsea Finn, Niccolo Fusai, Lachy Groom, Karol Hausman, Brian Ichter, et~al.
\newblock $pi_0$: A vision-language-action flow model for general robot control.
\newblock \emph{arXiv preprint arXiv:2410.24164}, 2024.

\bibitem[Blessing et~al.(2024)Blessing, Celik, Jia, Reuss, Li, Lioutikov, and Neumann]{blessing2024information}
Denis Blessing, Onur Celik, Xiaogang Jia, Moritz Reuss, Maximilian Li, Rudolf Lioutikov, and Gerhard Neumann.
\newblock Information maximizing curriculum: A curriculum-based approach for learning versatile skills.
\newblock \emph{Advances in Neural Information Processing Systems}, 36, 2024.

\bibitem[Bousmalis et~al.(2023)Bousmalis, Vezzani, Rao, Devin, Lee, Bauza, Davchev, Zhou, Gupta, Raju, et~al.]{bousmalis2023robocat}
Konstantinos Bousmalis, Giulia Vezzani, Dushyant Rao, Coline Devin, Alex~X Lee, Maria Bauza, Todor Davchev, Yuxiang Zhou, Agrim Gupta, Akhil Raju, et~al.
\newblock Robocat: A self-improving foundation agent for robotic manipulation.
\newblock \emph{arXiv preprint arXiv:2306.11706}, 2023.

\bibitem[Brohan et~al.(2022)Brohan, Brown, Carbajal, Chebotar, Dabis, Finn, Gopalakrishnan, Hausman, Herzog, Hsu, et~al.]{brohan2022rt}
Anthony Brohan, Noah Brown, Justice Carbajal, Yevgen Chebotar, Joseph Dabis, Chelsea Finn, Keerthana Gopalakrishnan, Karol Hausman, Alex Herzog, Jasmine Hsu, et~al.
\newblock Rt-1: Robotics transformer for real-world control at scale.
\newblock \emph{arXiv preprint arXiv:2212.06817}, 2022.

\bibitem[Brohan et~al.(2023)Brohan, Brown, Carbajal, Chebotar, Chen, Choromanski, Ding, Driess, Dubey, Finn, et~al.]{brohan2023rt}
Anthony Brohan, Noah Brown, Justice Carbajal, Yevgen Chebotar, Xi Chen, Krzysztof Choromanski, Tianli Ding, Danny Driess, Avinava Dubey, Chelsea Finn, et~al.
\newblock Rt-2: Vision-language-action models transfer web knowledge to robotic control.
\newblock \emph{arXiv preprint arXiv:2307.15818}, 2023.

\bibitem[Brooks et~al.(2023)Brooks, Holynski, and Efros]{brooks2023instructpix2pix}
Tim Brooks, Aleksander Holynski, and Alexei~A Efros.
\newblock Instructpix2pix: Learning to follow image editing instructions.
\newblock In \emph{Proceedings of the IEEE/CVF conference on computer vision and pattern recognition}, pages 18392--18402, 2023.

\bibitem[Brooks et~al.(2024)Brooks, Peebles, Holmes, DePue, Guo, Jing, Schnurr, Taylor, Luhman, Luhman, Ng, Wang, and Ramesh]{videoworldsimulators2024}
Tim Brooks, Bill Peebles, Connor Holmes, Will DePue, Yufei Guo, Li Jing, David Schnurr, Joe Taylor, Troy Luhman, Eric Luhman, Clarence Ng, Ricky Wang, and Aditya Ramesh.
\newblock Video generation models as world simulators.
\newblock 2024.

\bibitem[Cao et~al.(2024)Cao, Zhang, Sun, Wang, Cheng, Li, Ma, Shao, Zhao, Han, et~al.]{cao2024mamba}
Jiahang Cao, Qiang Zhang, Jingkai Sun, Jiaxu Wang, Hao Cheng, Yulin Li, Jun Ma, Yecheng Shao, Wen Zhao, Gang Han, et~al.
\newblock Mamba policy: Towards efficient 3d diffusion policy with hybrid selective state models.
\newblock \emph{arXiv preprint arXiv:2409.07163}, 2024.

\bibitem[Cheang et~al.(2024)Cheang, Chen, Jing, Kong, Li, Li, Liu, Wu, Xu, Yang, et~al.]{cheang2024gr}
Chi-Lam Cheang, Guangzeng Chen, Ya Jing, Tao Kong, Hang Li, Yifeng Li, Yuxiao Liu, Hongtao Wu, Jiafeng Xu, Yichu Yang, et~al.
\newblock Gr-2: A generative video-language-action model with web-scale knowledge for robot manipulation.
\newblock \emph{arXiv preprint arXiv:2410.06158}, 2024.

\bibitem[Chen et~al.(2024{\natexlab{a}})Chen, Monso, Du, Simchowitz, Tedrake, and Sitzmann]{chen2024diffusion}
Boyuan Chen, Diego~Marti Monso, Yilun Du, Max Simchowitz, Russ Tedrake, and Vincent Sitzmann.
\newblock Diffusion forcing: Next-token prediction meets full-sequence diffusion.
\newblock \emph{arXiv preprint arXiv:2407.01392}, 2024{\natexlab{a}}.

\bibitem[Chen et~al.(2023)Chen, Bahl, and Pathak]{chen2023playfusion}
Lili Chen, Shikhar Bahl, and Deepak Pathak.
\newblock Playfusion: Skill acquisition via diffusion from language-annotated play.
\newblock In \emph{Conference on Robot Learning}, pages 2012--2029. PMLR, 2023.

\bibitem[Chen et~al.(2024{\natexlab{b}})Chen, Wu, Wang, Su, Chen, Xing, Zhong, Zhang, Zhu, Lu, et~al.]{chen2024internvl}
Zhe Chen, Jiannan Wu, Wenhai Wang, Weijie Su, Guo Chen, Sen Xing, Muyan Zhong, Qinglong Zhang, Xizhou Zhu, Lewei Lu, et~al.
\newblock Internvl: Scaling up vision foundation models and aligning for generic visual-linguistic tasks.
\newblock In \emph{Proceedings of the IEEE/CVF conference on computer vision and pattern recognition}, pages 24185--24198, 2024{\natexlab{b}}.

\bibitem[Chi et~al.(2023)Chi, Feng, Du, Xu, Cousineau, Burchfiel, and Song]{chi2023diffusion}
Cheng Chi, Siyuan Feng, Yilun Du, Zhenjia Xu, Eric Cousineau, Benjamin Burchfiel, and Shuran Song.
\newblock Diffusion policy: Visuomotor policy learning via action diffusion.
\newblock \emph{arXiv preprint arXiv:2303.04137}, 2023.

\bibitem[Dasari et~al.(2024)Dasari, Mees, Zhao, Srirama, and Levine]{dasari2024ingredients}
Sudeep Dasari, Oier Mees, Sebastian Zhao, Mohan~Kumar Srirama, and Sergey Levine.
\newblock The ingredients for robotic diffusion transformers.
\newblock \emph{arXiv preprint arXiv:2410.10088}, 2024.

\bibitem[Dhariwal and Nichol(2021)]{dhariwal2021diffusion}
Prafulla Dhariwal and Alexander Nichol.
\newblock Diffusion models beat gans on image synthesis.
\newblock \emph{Advances in neural information processing systems}, 34:\penalty0 8780--8794, 2021.

\bibitem[Driess et~al.(2023)Driess, Xia, Sajjadi, Lynch, Chowdhery, Ichter, Wahid, Tompson, Vuong, Yu, et~al.]{driess2023palm}
Danny Driess, Fei Xia, Mehdi~SM Sajjadi, Corey Lynch, Aakanksha Chowdhery, Brian Ichter, Ayzaan Wahid, Jonathan Tompson, Quan Vuong, Tianhe Yu, et~al.
\newblock Palm-e: An embodied multimodal language model.
\newblock \emph{arXiv preprint arXiv:2303.03378}, 2023.

\bibitem[Fang et~al.(2023)Fang, Fang, Tang, Liu, Wang, Zhu, and Lu]{fang2023rh20t}
Hao-Shu Fang, Hongjie Fang, Zhenyu Tang, Jirong Liu, Junbo Wang, Haoyi Zhu, and Cewu Lu.
\newblock Rh20t: A robotic dataset for learning diverse skills in one-shot.
\newblock In \emph{RSS 2023 Workshop on Learning for Task and Motion Planning}, 2023.

\bibitem[Gu et~al.(2023)Gu, Xiang, Li, Ling, Liu, Mu, Tang, Tao, Wei, Yao, et~al.]{gu2023ManiSkill2}
Jiayuan Gu, Fanbo Xiang, Xuanlin Li, Zhan Ling, Xiqiang Liu, Tongzhou Mu, Yihe Tang, Stone Tao, Xinyue Wei, Yunchao Yao, et~al.
\newblock Maniskill2: A unified benchmark for generalizable manipulation skills.
\newblock \emph{arXiv preprint arXiv:2302.04659}, 2023.

\bibitem[Ha et~al.(2023)Ha, Florence, and Song]{ha2023scaling}
Huy Ha, Pete Florence, and Shuran Song.
\newblock Scaling up and distilling down: Language-guided robot skill acquisition.
\newblock In \emph{Conference on Robot Learning}, pages 3766--3777. PMLR, 2023.

\bibitem[Hatch et~al.(2024)Hatch, Balakrishna, Mees, Nair, Park, Wulfe, Itkina, Eysenbach, Levine, Kollar, et~al.]{hatch2024ghil}
Kyle~B Hatch, Ashwin Balakrishna, Oier Mees, Suraj Nair, Seohong Park, Blake Wulfe, Masha Itkina, Benjamin Eysenbach, Sergey Levine, Thomas Kollar, et~al.
\newblock Ghil-glue: Hierarchical control with filtered subgoal images.
\newblock \emph{arXiv preprint arXiv:2410.20018}, 2024.

\bibitem[Ho et~al.(2020)Ho, Jain, and Abbeel]{ho2020denoising}
Jonathan Ho, Ajay Jain, and Pieter Abbeel.
\newblock Denoising diffusion probabilistic models.
\newblock \emph{Advances in neural information processing systems}, 33:\penalty0 6840--6851, 2020.

\bibitem[Hu et~al.(2021)Hu, Shen, Wallis, Allen-Zhu, Li, Wang, Wang, and Chen]{hu2021lora}
Edward~J Hu, Yelong Shen, Phillip Wallis, Zeyuan Allen-Zhu, Yuanzhi Li, Shean Wang, Lu Wang, and Weizhu Chen.
\newblock Lora: Low-rank adaptation of large language models.
\newblock \emph{arXiv preprint arXiv:2106.09685}, 2021.

\bibitem[Huang et~al.(2023)Huang, Yong, Ma, Linghu, Li, Wang, Li, Zhu, Jia, and Huang]{huang2023embodied}
Jiangyong Huang, Silong Yong, Xiaojian Ma, Xiongkun Linghu, Puhao Li, Yan Wang, Qing Li, Song-Chun Zhu, Baoxiong Jia, and Siyuan Huang.
\newblock An embodied generalist agent in 3d world.
\newblock \emph{arXiv preprint arXiv:2311.12871}, 2023.

\bibitem[Huang et~al.(2025)Huang, Chen, Zhou, Chen, Jiang, Hu, Gao, Li, Yao, and Ren]{huang2025enerverse}
Siyuan Huang, Liliang Chen, Pengfei Zhou, Shengcong Chen, Zhengkai Jiang, Yue Hu, Peng Gao, Hongsheng Li, Maoqing Yao, and Guanghui Ren.
\newblock Enerverse: Envisioning embodied future space for robotics manipulation.
\newblock \emph{arXiv preprint arXiv:2501.01895}, 2025.

\bibitem[Karamcheti et~al.(2023)Karamcheti, Nair, Chen, Kollar, Finn, Sadigh, and Liang]{karamcheti2023language}
Siddharth Karamcheti, Suraj Nair, Annie~S Chen, Thomas Kollar, Chelsea Finn, Dorsa Sadigh, and Percy Liang.
\newblock Language-driven representation learning for robotics.
\newblock \emph{arXiv preprint arXiv:2302.12766}, 2023.

\bibitem[Ke et~al.(2024)Ke, Gkanatsios, and Fragkiadaki]{ke20243d}
Tsung-Wei Ke, Nikolaos Gkanatsios, and Katerina Fragkiadaki.
\newblock 3d diffuser actor: Policy diffusion with 3d scene representations.
\newblock \emph{arXiv preprint arXiv:2402.10885}, 2024.

\bibitem[Khazatsky et~al.(2024)Khazatsky, Pertsch, Nair, Balakrishna, Dasari, Karamcheti, Nasiriany, Srirama, Chen, Ellis, et~al.]{khazatsky2024droid}
Alexander Khazatsky, Karl Pertsch, Suraj Nair, Ashwin Balakrishna, Sudeep Dasari, Siddharth Karamcheti, Soroush Nasiriany, Mohan~Kumar Srirama, Lawrence~Yunliang Chen, Kirsty Ellis, et~al.
\newblock Droid: A large-scale in-the-wild robot manipulation dataset.
\newblock \emph{arXiv preprint arXiv:2403.12945}, 2024.

\bibitem[Kim et~al.(2024)Kim, Pertsch, Karamcheti, Xiao, Balakrishna, Nair, Rafailov, Foster, Lam, Sanketi, et~al.]{kim2024openvla}
Moo~Jin Kim, Karl Pertsch, Siddharth Karamcheti, Ted Xiao, Ashwin Balakrishna, Suraj Nair, Rafael Rafailov, Ethan Foster, Grace Lam, Pannag Sanketi, et~al.
\newblock Openvla: An open-source vision-language-action model.
\newblock \emph{arXiv preprint arXiv:2406.09246}, 2024.

\bibitem[Li et~al.(2023{\natexlab{a}})Li, Li, Savarese, and Hoi]{li2023blip}
Junnan Li, Dongxu Li, Silvio Savarese, and Steven Hoi.
\newblock Blip-2: Bootstrapping language-image pre-training with frozen image encoders and large language models.
\newblock In \emph{International conference on machine learning}, pages 19730--19742. PMLR, 2023{\natexlab{a}}.

\bibitem[Li et~al.(2025)Li, Wu, Huang, Cheang, Wang, and Kong]{li2025gr}
Peiyan Li, Hongtao Wu, Yan Huang, Chilam Cheang, Liang Wang, and Tao Kong.
\newblock Gr-mg: Leveraging partially-annotated data via multi-modal goal-conditioned policy.
\newblock \emph{IEEE Robotics and Automation Letters}, 2025.

\bibitem[Li et~al.(2024{\natexlab{a}})Li, Tian, Li, Deng, and He]{li2024autoregressive}
Tianhong Li, Yonglong Tian, He Li, Mingyang Deng, and Kaiming He.
\newblock Autoregressive image generation without vector quantization.
\newblock \emph{arXiv preprint arXiv:2406.11838}, 2024{\natexlab{a}}.

\bibitem[Li et~al.(2023{\natexlab{b}})Li, Liu, Zhang, Yu, Xu, Wu, Cheang, Jing, Zhang, Liu, et~al.]{li2023vision}
Xinghang Li, Minghuan Liu, Hanbo Zhang, Cunjun Yu, Jie Xu, Hongtao Wu, Chilam Cheang, Ya Jing, Weinan Zhang, Huaping Liu, et~al.
\newblock Vision-language foundation models as effective robot imitators.
\newblock \emph{arXiv preprint arXiv:2311.01378}, 2023{\natexlab{b}}.

\bibitem[Li et~al.(2024{\natexlab{b}})Li, Hsu, Gu, Pertsch, Mees, Walke, Fu, Lunawat, Sieh, Kirmani, Levine, Wu, Finn, Su, Vuong, and Xiao]{li24simpler}
Xuanlin Li, Kyle Hsu, Jiayuan Gu, Karl Pertsch, Oier Mees, Homer~Rich Walke, Chuyuan Fu, Ishikaa Lunawat, Isabel Sieh, Sean Kirmani, Sergey Levine, Jiajun Wu, Chelsea Finn, Hao Su, Quan Vuong, and Ted Xiao.
\newblock Evaluating real-world robot manipulation policies in simulation.
\newblock \emph{arXiv preprint arXiv:2405.05941}, 2024{\natexlab{b}}.

\bibitem[Li et~al.(2024{\natexlab{c}})Li, Li, Liu, Wang, Liu, Kang, Ma, Kong, Zhang, and Liu]{li2024towards}
Xinghang Li, Peiyan Li, Minghuan Liu, Dong Wang, Jirong Liu, Bingyi Kang, Xiao Ma, Tao Kong, Hanbo Zhang, and Huaping Liu.
\newblock Towards generalist robot policies: What matters in building vision-language-action models.
\newblock \emph{arXiv preprint arXiv:2412.14058}, 2024{\natexlab{c}}.

\bibitem[Liang et~al.(2024)Liang, Mu, Ma, Tomizuka, Ding, and Luo]{liang2024skilldiffuser}
Zhixuan Liang, Yao Mu, Hengbo Ma, Masayoshi Tomizuka, Mingyu Ding, and Ping Luo.
\newblock Skilldiffuser: Interpretable hierarchical planning via skill abstractions in diffusion-based task execution.
\newblock In \emph{Proceedings of the IEEE/CVF Conference on Computer Vision and Pattern Recognition}, pages 16467--16476, 2024.

\bibitem[Liu et~al.(2024{\natexlab{a}})Liu, Zhu, Gao, Feng, Liu, Zhu, and Stone]{liu2024libero}
Bo Liu, Yifeng Zhu, Chongkai Gao, Yihao Feng, Qiang Liu, Yuke Zhu, and Peter Stone.
\newblock Libero: Benchmarking knowledge transfer for lifelong robot learning.
\newblock \emph{Advances in Neural Information Processing Systems}, 36, 2024{\natexlab{a}}.

\bibitem[Liu et~al.(2023)Liu, Li, Wu, and Lee]{liu2023llava}
Haotian Liu, Chunyuan Li, Qingyang Wu, and Yong~Jae Lee.
\newblock Visual instruction tuning, 2023.

\bibitem[Liu et~al.(2024{\natexlab{b}})Liu, Wu, Li, Tan, Chen, Wang, Xu, Su, and Zhu]{liu2024rdt}
Songming Liu, Lingxuan Wu, Bangguo Li, Hengkai Tan, Huayu Chen, Zhengyi Wang, Ke Xu, Hang Su, and Jun Zhu.
\newblock Rdt-1b: a diffusion foundation model for bimanual manipulation.
\newblock \emph{arXiv preprint arXiv:2410.07864}, 2024{\natexlab{b}}.

\bibitem[Loshchilov(2017)]{loshchilov2017decoupled}
I Loshchilov.
\newblock Decoupled weight decay regularization.
\newblock \emph{arXiv preprint arXiv:1711.05101}, 2017.

\bibitem[Lynch and Sermanet(2020)]{lynch2020language}
Corey Lynch and Pierre Sermanet.
\newblock Language conditioned imitation learning over unstructured data.
\newblock \emph{arXiv preprint arXiv:2005.07648}, 2020.

\bibitem[Lynch et~al.(2020)Lynch, Khansari, Xiao, Kumar, Tompson, Levine, and Sermanet]{lynch2020learning}
Corey Lynch, Mohi Khansari, Ted Xiao, Vikash Kumar, Jonathan Tompson, Sergey Levine, and Pierre Sermanet.
\newblock Learning latent plans from play.
\newblock In \emph{Conference on robot learning}, pages 1113--1132. PMLR, 2020.

\bibitem[Mees et~al.(2022{\natexlab{a}})Mees, Hermann, and Burgard]{mees2022matters}
Oier Mees, Lukas Hermann, and Wolfram Burgard.
\newblock What matters in language conditioned robotic imitation learning over unstructured data.
\newblock \emph{IEEE Robotics and Automation Letters}, 7\penalty0 (4):\penalty0 11205--11212, 2022{\natexlab{a}}.

\bibitem[Mees et~al.(2022{\natexlab{b}})Mees, Hermann, Rosete-Beas, and Burgard]{mees2022calvin}
Oier Mees, Lukas Hermann, Erick Rosete-Beas, and Wolfram Burgard.
\newblock Calvin: A benchmark for language-conditioned policy learning for long-horizon robot manipulation tasks.
\newblock \emph{IEEE Robotics and Automation Letters}, 7\penalty0 (3):\penalty0 7327--7334, 2022{\natexlab{b}}.

\bibitem[Mu et~al.(2021)Mu, Ling, Xiang, Yang, Li, Tao, Huang, Jia, and Su]{mu2021maniskill}
Tongzhou Mu, Zhan Ling, Fanbo Xiang, Derek Yang, Xuanlin Li, Stone Tao, Zhiao Huang, Zhiwei Jia, and Hao Su.
\newblock Maniskill: Generalizable manipulation skill benchmark with large-scale demonstrations.
\newblock \emph{arXiv preprint arXiv:2107.14483}, 2021.

\bibitem[Myers et~al.(2023)Myers, He, Fang, Walke, Hansen-Estruch, Cheng, Jalobeanu, Kolobov, Dragan, and Levine]{myers2023goal}
Vivek Myers, Andre~Wang He, Kuan Fang, Homer~Rich Walke, Philippe Hansen-Estruch, Ching-An Cheng, Mihai Jalobeanu, Andrey Kolobov, Anca Dragan, and Sergey Levine.
\newblock Goal representations for instruction following: A semi-supervised language interface to control.
\newblock In \emph{Conference on Robot Learning}, pages 3894--3908. PMLR, 2023.

\bibitem[OpenAI(2021)]{openai_dall_e}
OpenAI.
\newblock Dall·e: Creating images from text, 2021.

\bibitem[OpenAI(2022)]{openai_chatgpt}
OpenAI.
\newblock Chatgpt, 2022.

\bibitem[OpenAI(2023)]{openai_gpt4v}
OpenAI.
\newblock Gpt-4: Multimodal capabilities, 2023.

\bibitem[Oquab et~al.(2023)Oquab, Darcet, Moutakanni, Vo, Szafraniec, Khalidov, Fernandez, Haziza, Massa, El-Nouby, et~al.]{oquab2023dinov2}
Maxime Oquab, Timoth{\'e}e Darcet, Th{\'e}o Moutakanni, Huy Vo, Marc Szafraniec, Vasil Khalidov, Pierre Fernandez, Daniel Haziza, Francisco Massa, Alaaeldin El-Nouby, et~al.
\newblock Dinov2: Learning robust visual features without supervision.
\newblock \emph{arXiv preprint arXiv:2304.07193}, 2023.

\bibitem[Padalkar et~al.(2023)Padalkar, Pooley, Jain, Bewley, Herzog, Irpan, Khazatsky, Rai, Singh, Brohan, et~al.]{padalkar2023open}
Abhishek Padalkar, Acorn Pooley, Ajinkya Jain, Alex Bewley, Alex Herzog, Alex Irpan, Alexander Khazatsky, Anant Rai, Anikait Singh, Anthony Brohan, et~al.
\newblock Open x-embodiment: Robotic learning datasets and rt-x models.
\newblock \emph{arXiv preprint arXiv:2310.08864}, 2023.

\bibitem[Pearce et~al.(2023)Pearce, Rashid, Kanervisto, Bignell, Sun, Georgescu, Macua, Tan, Momennejad, Hofmann, et~al.]{pearce2023imitating}
Tim Pearce, Tabish Rashid, Anssi Kanervisto, Dave Bignell, Mingfei Sun, Raluca Georgescu, Sergio~Valcarcel Macua, Shan~Zheng Tan, Ida Momennejad, Katja Hofmann, et~al.
\newblock Imitating human behaviour with diffusion models.
\newblock \emph{arXiv preprint arXiv:2301.10677}, 2023.

\bibitem[Peebles and Xie(2023)]{peebles2023scalable}
William Peebles and Saining Xie.
\newblock Scalable diffusion models with transformers.
\newblock In \emph{Proceedings of the IEEE/CVF International Conference on Computer Vision}, pages 4195--4205, 2023.

\bibitem[Perez et~al.(2018)Perez, Strub, De~Vries, Dumoulin, and Courville]{perez2018film}
Ethan Perez, Florian Strub, Harm De~Vries, Vincent Dumoulin, and Aaron Courville.
\newblock Film: Visual reasoning with a general conditioning layer.
\newblock In \emph{Proceedings of the AAAI conference on artificial intelligence}, 2018.

\bibitem[Qu et~al.(2025)Qu, Song, Chen, Yao, Ye, Ding, Wang, Gu, Zhao, Wang, et~al.]{qu2025spatialvla}
Delin Qu, Haoming Song, Qizhi Chen, Yuanqi Yao, Xinyi Ye, Yan Ding, Zhigang Wang, JiaYuan Gu, Bin Zhao, Dong Wang, et~al.
\newblock Spatialvla: Exploring spatial representations for visual-language-action model.
\newblock \emph{arXiv preprint arXiv:2501.15830}, 2025.

\bibitem[Radford et~al.(2021)Radford, Kim, Hallacy, Ramesh, Goh, Agarwal, Sastry, Askell, Mishkin, Clark, et~al.]{radford2021learning}
Alec Radford, Jong~Wook Kim, Chris Hallacy, Aditya Ramesh, Gabriel Goh, Sandhini Agarwal, Girish Sastry, Amanda Askell, Pamela Mishkin, Jack Clark, et~al.
\newblock Learning transferable visual models from natural language supervision.
\newblock In \emph{International conference on machine learning}, pages 8748--8763. PMLR, 2021.

\bibitem[Reuss et~al.(2023)Reuss, Li, Jia, and Lioutikov]{reuss2023goal}
Moritz Reuss, Maximilian Li, Xiaogang Jia, and Rudolf Lioutikov.
\newblock Goal-conditioned imitation learning using score-based diffusion policies.
\newblock \emph{arXiv preprint arXiv:2304.02532}, 2023.

\bibitem[Reuss et~al.(2024)Reuss, Ya{\u{g}}murlu, Wenzel, and Lioutikov]{reuss2024multimodal}
Moritz Reuss, {\"O}mer~Erdin{\c{c}} Ya{\u{g}}murlu, Fabian Wenzel, and Rudolf Lioutikov.
\newblock Multimodal diffusion transformer: Learning versatile behavior from multimodal goals.
\newblock In \emph{First Workshop on Vision-Language Models for Navigation and Manipulation at ICRA 2024}, 2024.

\bibitem[Rombach et~al.(2021)Rombach, Blattmann, Lorenz, Esser, and Schmidt]{rombach2021stablediffusion}
Robin Rombach, Andreas Blattmann, Dominik Lorenz, Patrick Esser, and Tim Schmidt.
\newblock Stable diffusion: High-resolution image synthesis with latent diffusion models, 2021.

\bibitem[Rombach et~al.(2022)Rombach, Blattmann, Lorenz, Esser, and Ommer]{rombach2022high}
Robin Rombach, Andreas Blattmann, Dominik Lorenz, Patrick Esser, and Bj{\"o}rn Ommer.
\newblock High-resolution image synthesis with latent diffusion models.
\newblock In \emph{Proceedings of the IEEE/CVF conference on computer vision and pattern recognition}, pages 10684--10695, 2022.

\bibitem[Scheikl et~al.(2024)Scheikl, Schreiber, Haas, Freymuth, Neumann, Lioutikov, and Mathis-Ullrich]{scheikl2024movement}
Paul~Maria Scheikl, Nicolas Schreiber, Christoph Haas, Niklas Freymuth, Gerhard Neumann, Rudolf Lioutikov, and Franziska Mathis-Ullrich.
\newblock Movement primitive diffusion: Learning gentle robotic manipulation of deformable objects.
\newblock \emph{IEEE Robotics and Automation Letters}, 2024.

\bibitem[Shah et~al.(2023{\natexlab{a}})Shah, Sridhar, Bhorkar, Hirose, and Levine]{shah2023gnm}
Dhruv Shah, Ajay Sridhar, Arjun Bhorkar, Noriaki Hirose, and Sergey Levine.
\newblock Gnm: A general navigation model to drive any robot.
\newblock In \emph{2023 IEEE International Conference on Robotics and Automation (ICRA)}, pages 7226--7233. IEEE, 2023{\natexlab{a}}.

\bibitem[Shah et~al.(2023{\natexlab{b}})Shah, Sridhar, Dashora, Stachowicz, Black, Hirose, and Levine]{shah2023vint}
Dhruv Shah, Ajay Sridhar, Nitish Dashora, Kyle Stachowicz, Kevin Black, Noriaki Hirose, and Sergey Levine.
\newblock Vint: A foundation model for visual navigation.
\newblock \emph{arXiv preprint arXiv:2306.14846}, 2023{\natexlab{b}}.

\bibitem[Shah et~al.(2023{\natexlab{c}})Shah, Mart{\'\i}n-Mart{\'\i}n, and Zhu]{shah2023mutex}
Rutav Shah, Roberto Mart{\'\i}n-Mart{\'\i}n, and Yuke Zhu.
\newblock Mutex: Learning unified policies from multimodal task specifications.
\newblock \emph{arXiv preprint arXiv:2309.14320}, 2023{\natexlab{c}}.

\bibitem[Shridhar et~al.(2023)Shridhar, Manuelli, and Fox]{shridhar2023perceiver}
Mohit Shridhar, Lucas Manuelli, and Dieter Fox.
\newblock Perceiver-actor: A multi-task transformer for robotic manipulation.
\newblock In \emph{Conference on Robot Learning}, pages 785--799. PMLR, 2023.

\bibitem[Song et~al.(2020)Song, Meng, and Ermon]{song2020denoising}
Jiaming Song, Chenlin Meng, and Stefano Ermon.
\newblock Denoising diffusion implicit models.
\newblock \emph{arXiv preprint arXiv:2010.02502}, 2020.

\bibitem[Song et~al.(2023)Song, Dhariwal, Chen, and Sutskever]{song2023consistency}
Yang Song, Prafulla Dhariwal, Mark Chen, and Ilya Sutskever.
\newblock Consistency models.
\newblock 2023.

\bibitem[Sridhar et~al.(2024)Sridhar, Shah, Glossop, and Levine]{sridhar2024nomad}
Ajay Sridhar, Dhruv Shah, Catherine Glossop, and Sergey Levine.
\newblock Nomad: Goal masked diffusion policies for navigation and exploration.
\newblock In \emph{2024 IEEE International Conference on Robotics and Automation (ICRA)}, pages 63--70. IEEE, 2024.

\bibitem[Team et~al.(2024)Team, Ghosh, Walke, Pertsch, Black, Mees, Dasari, Hejna, Kreiman, Xu, et~al.]{team2024octo}
Octo~Model Team, Dibya Ghosh, Homer Walke, Karl Pertsch, Kevin Black, Oier Mees, Sudeep Dasari, Joey Hejna, Tobias Kreiman, Charles Xu, et~al.
\newblock Octo: An open-source generalist robot policy.
\newblock \emph{arXiv preprint arXiv:2405.12213}, 2024.

\bibitem[Tian et~al.(2024)Tian, Yang, Zeng, Wang, Lin, Dong, and Pang]{tian2024predictive}
Yang Tian, Sizhe Yang, Jia Zeng, Ping Wang, Dahua Lin, Hao Dong, and Jiangmiao Pang.
\newblock Predictive inverse dynamics models are scalable learners for robotic manipulation.
\newblock \emph{arXiv preprint arXiv:2412.15109}, 2024.

\bibitem[Wang et~al.(2024{\natexlab{a}})Wang, Zhang, Huo, Tian, Zhang, Xie, Xu, Ji, Zhan, Ding, et~al.]{wang2024sparse}
Yixiao Wang, Yifei Zhang, Mingxiao Huo, Ran Tian, Xiang Zhang, Yichen Xie, Chenfeng Xu, Pengliang Ji, Wei Zhan, Mingyu Ding, et~al.
\newblock Sparse diffusion policy: A sparse, reusable, and flexible policy for robot learning.
\newblock \emph{arXiv preprint arXiv:2407.01531}, 2024{\natexlab{a}}.

\bibitem[Wang et~al.(2024{\natexlab{b}})Wang, Li, Mandlekar, Xu, Fan, Narang, Fan, Zhu, Balaji, Zhou, et~al.]{wang2024one}
Zhendong Wang, Zhaoshuo Li, Ajay Mandlekar, Zhenjia Xu, Jiaojiao Fan, Yashraj Narang, Linxi Fan, Yuke Zhu, Yogesh Balaji, Mingyuan Zhou, et~al.
\newblock One-step diffusion policy: Fast visuomotor policies via diffusion distillation.
\newblock \emph{arXiv preprint arXiv:2410.21257}, 2024{\natexlab{b}}.

\bibitem[Wen et~al.(2024)Wen, Zhu, Zhu, Tang, Li, Zhou, Li, Liu, Peng, Shen, et~al.]{wen2024diffusion}
Junjie Wen, Minjie Zhu, Yichen Zhu, Zhibin Tang, Jinming Li, Zhongyi Zhou, Chengmeng Li, Xiaoyu Liu, Yaxin Peng, Chaomin Shen, et~al.
\newblock Diffusion-vla: Scaling robot foundation models via unified diffusion and autoregression.
\newblock \emph{arXiv preprint arXiv:2412.03293}, 2024.

\bibitem[Wen et~al.(2025)Wen, Zhu, Li, Tang, Shen, and Feng]{wen2025dexvla}
Junjie Wen, Yichen Zhu, Jinming Li, Zhibin Tang, Chaomin Shen, and Feifei Feng.
\newblock Dexvla: Vision-language model with plug-in diffusion expert for general robot control.
\newblock \emph{arXiv preprint arXiv:2502.05855}, 2025.

\bibitem[Wu et~al.(2023)Wu, Jing, Cheang, Chen, Xu, Li, Liu, Li, and Kong]{wu2023unleashing}
Hongtao Wu, Ya Jing, Chilam Cheang, Guangzeng Chen, Jiafeng Xu, Xinghang Li, Minghuan Liu, Hang Li, and Tao Kong.
\newblock Unleashing large-scale video generative pre-training for visual robot manipulation.
\newblock \emph{arXiv preprint arXiv:2312.13139}, 2023.

\bibitem[Xian et~al.(2023)Xian, Gkanatsios, Gervet, and Fragkiadaki]{xian2023unifying}
Zhou Xian, Nikolaos Gkanatsios, Theophile Gervet, and Katerina Fragkiadaki.
\newblock Unifying diffusion models with action detection transformers for multi-task robotic manipulation.
\newblock In \emph{7th Annual Conference on Robot Learning}, page~5, 2023.

\bibitem[Xiao et~al.(2022)Xiao, Chan, Sermanet, Wahid, Brohan, Hausman, Levine, and Tompson]{xiao2022robotic}
Ted Xiao, Harris Chan, Pierre Sermanet, Ayzaan Wahid, Anthony Brohan, Karol Hausman, Sergey Levine, and Jonathan Tompson.
\newblock Robotic skill acquisition via instruction augmentation with vision-language models.
\newblock \emph{arXiv preprint arXiv:2211.11736}, 2022.

\bibitem[Xing et~al.(2024)Xing, Xia, Zhang, Chen, Yu, Liu, Liu, Wang, Shan, and Wong]{xing2024dynamicrafter}
Jinbo Xing, Menghan Xia, Yong Zhang, Haoxin Chen, Wangbo Yu, Hanyuan Liu, Gongye Liu, Xintao Wang, Ying Shan, and Tien-Tsin Wong.
\newblock Dynamicrafter: Animating open-domain images with video diffusion priors.
\newblock In \emph{European Conference on Computer Vision}, pages 399--417. Springer, 2024.

\bibitem[Yang et~al.(2024)Yang, Glossop, Bhorkar, Shah, Vuong, Finn, Sadigh, and Levine]{yang2024pushing}
Jonathan Yang, Catherine Glossop, Arjun Bhorkar, Dhruv Shah, Quan Vuong, Chelsea Finn, Dorsa Sadigh, and Sergey Levine.
\newblock Pushing the limits of cross-embodiment learning for manipulation and navigation.
\newblock \emph{arXiv preprint arXiv:2402.19432}, 2024.

\bibitem[Ze et~al.(2024)Ze, Zhang, Zhang, Hu, Wang, and Xu]{ze20243d}
Yanjie Ze, Gu Zhang, Kangning Zhang, Chenyuan Hu, Muhan Wang, and Huazhe Xu.
\newblock 3d diffusion policy: Generalizable visuomotor policy learning via simple 3d representations.
\newblock In \emph{ICRA 2024 Workshop on 3D Visual Representations for Robot Manipulation}, 2024.

\bibitem[Zhang et~al.(2022)Zhang, Lu, Wang, and Zhang]{zhang2022language}
Edwin Zhang, Yujie Lu, William Wang, and Amy Zhang.
\newblock Language control diffusion: Efficiently scaling through space, time, and tasks.
\newblock \emph{arXiv preprint arXiv:2210.15629}, 2022.

\bibitem[Zhao et~al.(2023)Zhao, Kumar, Levine, and Finn]{zhao2023learning}
Tony~Z Zhao, Vikash Kumar, Sergey Levine, and Chelsea Finn.
\newblock Learning fine-grained bimanual manipulation with low-cost hardware.
\newblock \emph{arXiv preprint arXiv:2304.13705}, 2023.

\end{thebibliography}
